\PassOptionsToPackage{sort}{natbib}
\documentclass[Afour,times,SageH]{sagej}

\usepackage{float}
\usepackage{xurl}
\usepackage{placeins}
\usepackage{makecell}
\usepackage{booktabs}

\usepackage{amsmath}
\usepackage{amssymb}
\usepackage{amsthm}
\usepackage{bm}

\usepackage{algorithm}
\usepackage{algpseudocode}

\usepackage{graphicx}
\usepackage[font=footnotesize]{caption}

\usepackage[table,xcdraw]{xcolor}

\usepackage{natbib}

\usepackage[hidelinks,bookmarksopen=true,bookmarksnumbered=true]{hyperref}

\newcommand{\vertindu}{\mathbf{u}}
\newcommand{\moi}{\mathbf{I}^\text{i}}
\newcommand{\timeind}[2]{\mathbf{#1}_\text{#2}}
\newcommand{\timevertind}[3]{\mathbf{#1}^{#2}_{#3}}
\newcommand{\materialp}{\bm{\alpha}}
\newcommand{\Massmatrix}{\mathbf{M}}
\newcommand{\DEFT}{DEFT}
\newcommand{\DEFTn}{DEFT }
\newcommand{\MaterialFrame}{\Gamma}
\DeclareMathOperator*{\argmin}{arg\,min}
\newcommand{\bcurvature}{\kappa\mathbf{b}^i}
\newcommand{\R}{\mathbb{R}}
\newcommand{\secref}[1]{Section~\ref{#1}}
\newcommand{\pred}[2]{\hat{\mathbf{#1}}_\text{#2}}
\newtheorem{defn}{Definition}
\newtheorem{thm}[defn]{Theorem}
\DeclareMathOperator{\Exp}{Exp}
\DeclareMathOperator{\Log}{Log}

\definecolor{init_config}{RGB}{30,30,255}
\definecolor{mid_config}{RGB}{238,232,0}
\definecolor{target_hole}{RGB}{254,81,13}
\definecolor{contribution}{cmyk}{0.85,0.1, 1, 0.1}

\def\volumeyear{2026}

\begin{document}

\runninghead{Chen et al.}

\title{DEFT: Differentiable Branched Discrete Elastic Rods for Modeling Furcated DLOs in Real-Time}

\author{Yizhou Chen\affilnum{1}, Felicia Sang\affilnum{1}$^*$, Shicheng Peng\affilnum{1}$^*$, Xiaoyue Wu\affilnum{1}$^*$, Yeheng Zong\affilnum{1},
Yuzhen Chen\affilnum{1}, Anran Li\affilnum{1}, Bohao Zhang\affilnum{1},
and Ramanarayan Vasudevan\affilnum{1}}

\affiliation{\affilnum{1}Department of Robotics, University of Michigan, Ann Arbor, MI, USA}

\corrauth{$^*$These authors contributed equally.\\
Yizhou Chen, Roahm Lab, Department of Robotics, University of Michigan,
2505 Hayward St, Ann Arbor, MI 48109, USA.}

\email{yizhouch@umich.edu}
\begin{abstract}
Autonomous wire harness assembly requires robots to manipulate complex branched cables with high precision and reliability.
A key challenge to automating this process is accurately predicting how branched cables deform during manipulation.
Although well-studied for single Deformable Linear Objects (DLOs), existing methods do not extend directly to Branched DLOs (BDLOs), whose junctions couple bending, twisting, and stretching across branches in ways that connecting multiple DLO models cannot capture.
To address this challenge, this paper introduces Differentiable discrete branched Elastic rods for modeling Furcated DLOs in real-Time (\DEFT), a framework that combines a differentiable physics-based model with learning to:
1) accurately model BDLO dynamics, including dynamic propagation at junctions and grasping in the middle of a BDLO,
2) achieve real-time inference, and
3) support planning for dexterous BDLO manipulation.
\DEFTn introduces momentum-conserving junction constraints to explicitly capture cross-branch coupling, combines analytical gradients with branch-parallel computation for efficient parameter learning and inference, and uses a residual Graph Neural Network to reduce errors accumulated during long-horizon predictions.
Experiments on six physical BDLOs with different topologies and material properties demonstrate formulation-level generalization: the same DEFT formulation and residual-GNN architecture are applied across all evaluated BDLOs without redesign, while the physical and learned parameters are fitted separately for each BDLO.
\DEFTn outperforms state-of-the-art physics-based and learning-based methods.
Over five-second recursive open-loop predictions, \DEFTn achieves RMSEs of \(0.93\)--\(2.14\)~cm across the evaluated BDLOs while supporting real-time inference at over \(100\)~Hz.
In real-world planning, \DEFTn achieves success rates of \(36/40\) for shape matching, \(40/40\) for parent-branch insertion, and \(16/20\) for child-branch insertion.
These results demonstrate \DEFT's ability to provide accurate, efficient, and planning-compatible BDLO predictions for autonomous wire-harness manipulation.
\end{abstract}


\maketitle

\section{Introduction}
\label{section:Introduction}
\begin{figure*}[t]
    \centering
    \includegraphics[width=1\textwidth]{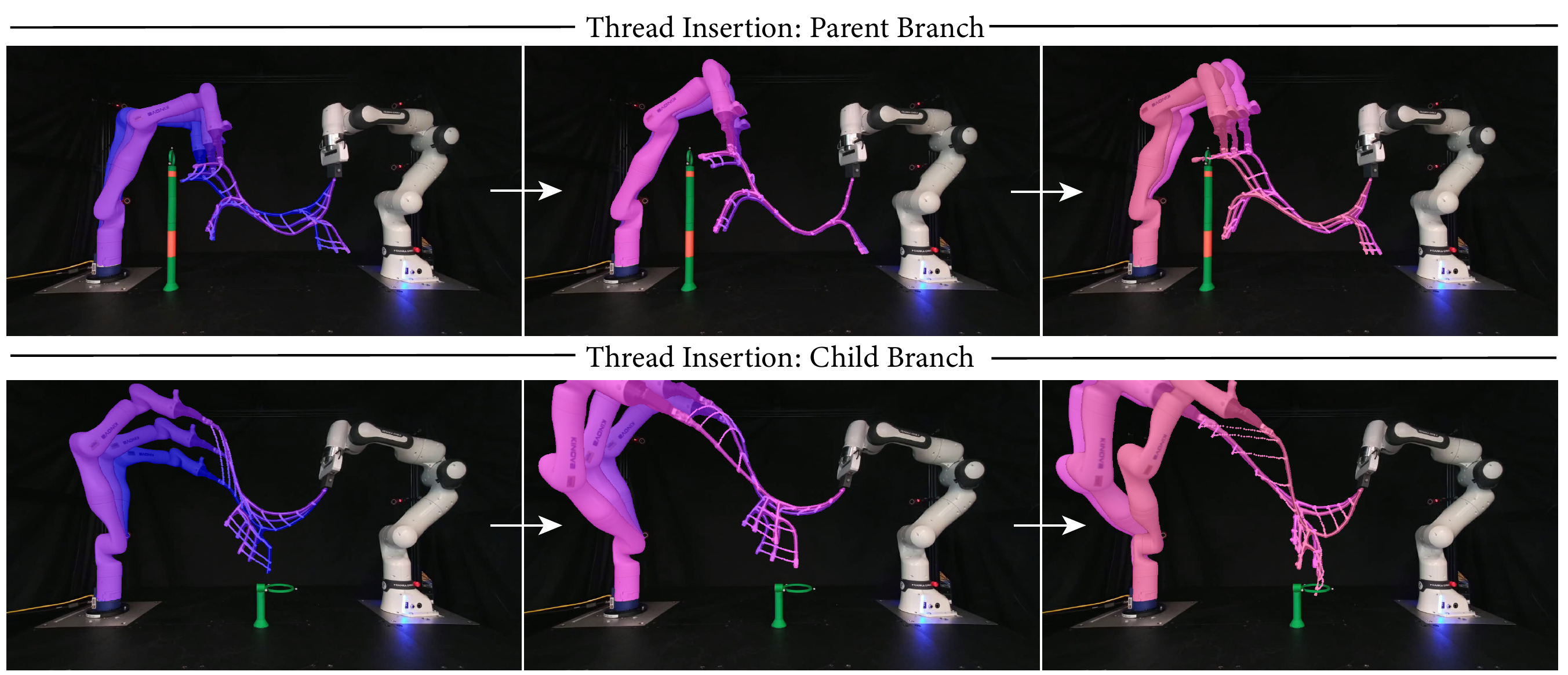}
    \caption{
    As this paper illustrates, DEFT can be used in concert with a motion planning algorithm to autonomously manipulate BDLOs.
    The figures above illustrate the ability of DEFT to autonomously perform wire insertion across different BDLO topologies and insertion configurations.
    Starting from a random \textcolor{init_config}{initial} configuration, the system guides the BDLO tip into the \textcolor{target_hole}{target hole}.
    The figures present two representative cases:
    \textbf{Top:} insertion of the main branch tip into the target hole;
    \textbf{Bottom:} insertion of a child branch tip in a BDLO with a different topology, while only the parent branch is grasped.
    Notably, DEFT recursively predicts the state of the wire throughout the process without relying on ground-truth or perception data at any stage.
    }
    \label{fig:first_page}
    \vspace{-0.6em}
\end{figure*}

The automation of wire harness assembly represents a critical challenge in manufacturing, particularly in the automotive and aerospace industries where complex, branched cable systems are essential components.
These assemblies require robots to manipulate Branched Deformable Linear Objects (BDLOs) with precision over extended periods (\(\geq5\)s)~\citep{deform_survey1, deform_survey2, saha2007manipulation}, a task that remains largely unautomated due to the objects' complex nonlinear dynamics.
Further complicating matters, practical assembly environments often involve occlusions, limiting perception systems' ability to estimate full BDLO configurations reliably~\citep{overview}.
Thus, robust assembly requires accurate BDLO models for reliable long-term open-loop predictions.
While accurate prediction of BDLO behavior is possible through computationally intensive models, real-time manipulation demands rapid computation.

Effective BDLO modeling requires three capabilities: real-time prediction, accurate 3D simulation, and efficient model parameter learning.
Real-time prediction is necessary for robotic manipulation, while 3D accuracy requires capturing bending, twisting, stretching, and dynamic coupling across branches.
Efficient parameter learning allows object-specific parameters to be identified for different materials and configurations without extensive manual tuning or data collection.
Achieving these capabilities simultaneously remains challenging: increasing physical fidelity often raises computational cost, while efficient data-driven models may require large datasets and generalize poorly across BDLO configurations.
We therefore compare existing DLO and BDLO modeling approaches according to these three capabilities.

\subsection{DLO Modeling}

Approaches to modeling DLOs have evolved from purely physics-based methods toward learning-based frameworks that balance the three aforementioned capabilities.

Physics-based approaches implement different mathematical frameworks to model physical behavior.
Mass-spring systems represent objects as networks of point masses connected by springs, computing forces and accelerations to simulate movement~\citep{LargeStepSimulation, diffcloth, Liu:2013:FSM}.
Position-Based Dynamics (PBD) directly manipulates vertex positions to satisfy physical constraints, avoiding expensive force calculations~\citep{directionalrigidity, diminishingridgidity, xpbd}.
Finite Element Methods (FEM) divide objects into discrete elements and solve partial differential equations to compute deformations~\citep{garcia2006optimized, sin2013vega, koessler2021efficient}.
Discrete Elastic Rods (DER) models represent objects as centerline curves with material frames, enabling explicit modeling of bending and twisting~\citep{diminishingridgidity, DER_manipulation1, DER_manipulation2, DER_manipulation3}.
Unfortunately, each of these methods has limitations: mass-spring systems describe the rigidity of wires by introducing large spring stiffness, which can make prediction numerically unstable; PBD models struggle with parameter learning because they assume that the system is quasi-static; FEM methods are computationally intensive; and conventional DER models lack differentiable expressions for their model parameters, making efficient parameter optimization challenging.

Learning-based modeling approaches use data-driven techniques to predict deformation behavior.
Methods such as Graph Neural Networks (GNNs) process vertex-level spatial information~\citep{GNN_baseline}, while Bi-LSTM models capture temporal dependencies in deformation sequences~\citep{bi-LSTM_baseline}.
These approaches enable real-time prediction through efficient computation and support straightforward parameter learning through standard optimization techniques.
However, they struggle with 3D accuracy, particularly in capturing the complex dynamic deformations characteristic of real-world manipulation.

Recent work has explored hybrid approaches that combine learning with physics-based modeling~\citep{mlphysicsprocess, physicslearn1, physicslearn2, physicslearn3, DEFORM, pinn1}.
This is exemplified by DEFORM~\citep{DEFORM}, which constructs a novel differentiable DER (DDER) model and combines it with a neural network-based residual-learning approach to account for numerical integration errors during prediction.
This hybrid approach achieves both real-time performance and 3D accuracy while enabling efficient parameter learning through end-to-end differentiability.
However, DEFORM lacks fundamental capabilities essential for accurate and efficient BDLO modeling.
Our experiments show that directly extending DEFORM to BDLOs reduces computational efficiency and fails to capture junction dynamics, leading to inaccurate predictions.

\subsection{BDLO Modeling}
\label{sebsection:BDLOs Modeling}

Neural network approaches such as Tree-LSTM~\citep{bi-LSTM_baseline, Tree-LSTM2} and GNNs~\citep{GNN_baseline, GCN} provide efficient real-time prediction and parameter learning, but our experiments show that these methods struggle to capture the complex 3D dynamic deformations that occur during manipulation tasks.
AdaptiGraph~\citep{AdaptiGraph} uses a fine-tuning strategy that leverages physics-based simulators: it randomizes material configurations across simulated dataset instances and subsequently fine-tunes the model on newly collected data.
However, without an accurate BDLO model, our experiments show that this approach, even when trained on datasets generated using finely stepped FEM simulations, underperforms DEFT.

Existing BDLO modeling approaches adapt various physical simulation frameworks to handle branched structures.
A straightforward approach is to rigidly align all connected branches with the perturbed branch, which sidesteps the numerical challenges associated with stiff dynamics but bypasses the underlying physics, leading to unrealistic predictions and discontinuous gradients~\citep{multibody2}.
Newtonian, FEM-based, and multibody methods model branches as connected elastic bodies with explicit junction interactions~\citep{Newtonian,FEM,LargeStepSimulation,FEM1,BDLO_FEM1,multibody,multibody2}.
These approaches can achieve high 3D accuracy but demand extensive parameter tuning and often require prohibitively small time steps, making them unsuitable for real-time applications.

Lagrangian-based, PBD, and projective-dynamics formulations~\citep{PBD_DER,bdlo_modeling1,originalDER,baseline1,projectivedynamics} stabilize stiff dynamics by casting each time step as a constrained minimization problem, enabling real-time performance.
More recent work extends this optimization perspective to gradient computation, showing that differentiation through the optimization solve can yield robust backpropagation even under stiff dynamics~\citep{diffpd, diffpd2, contactoptimization}.
However, these methods typically neglect dynamic effects and lack explicit constraints for modeling coupled stretching, bending, and twisting at junctions, making parameter learning and accurate BDLO simulation challenging~\citep{diffpd, diffpd2}.

Taken together, existing methods do not simultaneously provide accurate dynamic junction modeling, real-time prediction, and efficient gradient-based parameter learning.

A promising direction is to combine a differentiable BDLO physics model with a residual DNN to improve prediction accuracy and sample efficiency while supporting formulation-level generalization across different BDLO structures~\citep{PINN, mlphysicsprocess, physicslearn1, physicslearn2, physicslearn3, DEFORM, pinn1}.
However, such an integrated framework has not yet been developed for BDLOs, primarily due to three key challenges.
First, for forward simulation, there is no physics model that is both accurate and computationally efficient for BDLOs.
Existing formulations either sacrifice fidelity for speed or require prohibitively small time steps to remain stable.
Second, for backward gradient propagation, differentiability must be preserved when designing the physics model to enable efficient parameter optimization.
This requirement becomes particularly challenging under stiff junction dynamics~\citep{contactnets,physicslearn2}, where backpropagation through stiff modes can lead to unstable or ill-conditioned gradients.
Third, effectively learning BDLO-specific residual dynamics remains challenging because formulation-level generalization requires the same residual-learning architecture to accommodate diverse junction topologies and manipulation scenarios.

To address these challenges, this paper introduces
\textbf{D}ifferentiable Branched Discrete \textbf{E}lastic Rods for Modeling
\textbf{F}urcated DLOs in Real-\textbf{T}ime (\DEFT),
a novel differentiable physics-based framework that accurately predicts dynamic BDLO behavior over long time horizons with real-time inference capabilities.
Here, generalization across BDLOs refers to applying the same DEFT formulation and residual-GNN architecture without redesign; the physical and learned parameters are identified separately for each BDLO. Transferring fitted parameters to an unseen deformable object is a separate challenge for which prior work commonly employs model adaptation or online parameter estimation~\citep{adaptationdynamics,onlineadapt}.

The contributions of this paper are twofold:

\begin{enumerate}

\item
A novel differentiable BDLO model that captures dynamic propagation through wire-harness junctions and supports gradient backpropagation through a novel constraint-enforcement formulation.
This enables efficient identification of model parameters directly from experimental observations.

\item
An integrated computational and learning framework that combines a parallelized representation of BDLO dynamics for efficient real-time inference, analytical gradients for model parameter learning, and a specialized GNN for residual learning.
The same residual-learning architecture compensates for numerical integration errors during long-horizon prediction and provides formulation-level generalization across different BDLO topologies, while its parameters are trained separately for each BDLO.

\end{enumerate}

We validate \DEFTn through comprehensive real-world experiments across different BDLO topologies and manipulation scenarios.
The results demonstrate improved prediction accuracy, computational efficiency, sampling efficiency, and planning success compared with existing physics-based and learning-based approaches.
Figure~\ref{fig:first_page} illustrates \DEFTn used with a motion planner to autonomously insert BDLOs of different topologies into a target hole.
The framework is implemented in PyTorch for seamless integration with deep learning workflows. Based on the existing BDLO studies reviewed above, \DEFTn is the first framework to jointly provide quantitative evaluations of dynamic BDLO modeling and real-world robotic manipulation.
\section{Preliminaries}

\subsection{Deformable Linear Object (DLO) Model}
To model a DLO, we represent it as an indexed set of $n$ vertices at each time instance.
The ground truth location of vertex $i$ at time $t$ is denoted by $\mathbf{x}_t^i \in \mathbb{R}^3$, and the set of all $n$ vertices is denoted by $\mathbf{X}_t$.
A segment, or edge, in the DLO is the line between successive vertices, $\timevertind{e}{i}{t} = \timevertind{x}{i+1}{t} - \timevertind{x}{i}{t}$.
Let $\timeind{E}{t}$ correspond to the set of all edges at time $t$.
Let $\mathbf{M}_i \in \mathbb{R}^{3 \times 3}$ denote the mass matrix of vertex $i$.
The velocity of the vertices of the DLO is approximated by $\mathbf{V}_t = (\mathbf{X}_t - \mathbf{X}_{t-1})/{\Delta t}$, where $\Delta t$ is the time step between frames.
Note that $\mathbf{V}_t$ is an approximation of the actual velocities.
We distinguish between ground truth elements and predicted elements by using the circumflex symbol (e.g., $\mathbf{X}_t$ is the ground truth set of vertices at time $t$, and $\hat{\mathbf{X}}_t$ is the predicted set of vertices at time $t$).

\subsection{Modeling DLOs with DER}
\subsubsection{Discrete Elastic Rods (DER)}
DER models three distinct deformation modes of DLOs: bending, twisting, and stretching.
DER theory models DLOs using two key families of coordinate frames.
First, it introduces Bishop Frames, which provide a twist-free reference state along the centerline of the rod.
The Bishop Frame only rotates to follow the curve's geometry, providing a relaxed baseline configuration for the DLO.
Second, the theory employs Material Frames, which describe the actual physical deformation of the rod.
Material Frames can be expressed relative to the Bishop Frames through a single angle whose rate of change represents the physical twist of the DLO.
To describe the twist of the DLO, DER theory assumes that the angle of the Material Frame with respect to the Bishop Frame is one that minimizes the potential energy of the DLO.
We adopt this shear-free formulation because wire harnesses are thin, flexible cables for which the Kirchhoff assumption of negligible shear is well-justified by classical rod theory; extensions to shear-inclusive formulations~\citep{shearDER} are discussed in Section \ref{sec:discussion}.
We briefly summarize DER theory in this section, but a longer introduction can be found here
\mbox{\protect\citep{originalDER}}

\textbf{Bishop Frame}:
The Bishop Frame at edge $i$ is made up of three axes $\{\mathbf{t}^i, \mathbf{b}_1^i, \mathbf{b}_2^i\}$.
We next formally define each axis of the Bishop Frame.
First, $\mathbf{t}^i \in \mathbb{R}^3$ is the unit tangent along edge $i$ which is defined as $\mathbf{t}^i = (\timevertind{x}{i+1}{}-\timevertind{x}{i}{})/||\timevertind{x}{i+1}{}-\timevertind{x}{i}{}||_2$.
The vector $\mathbf{b}_1^i \in \mathbb{R}^{3}$ is defined iteratively by applying a rotation matrix to the previous Bishop frame: $\mathbf{b}_1^i = \mathbf{R}^i \cdot \mathbf{b}_1^{i-1}$.
The rotation matrix $\mathbf{R}^i \in \mathbb{R}^{3 \times 3}$ satisfies the following conditions
\begin{equation}
    \label{eq:rotation1}
    \mathbf{t}^i = \mathbf{R}^i \cdot \mathbf{t}^{i-1}
\end{equation}
\begin{equation}
    \label{eq:rotation2}
     \mathbf{t}^{i-1} \times \mathbf{t}^i = \mathbf{R}^i \cdot (\mathbf{t}^{i-1} \times \mathbf{t}^i).
\end{equation}
The first condition ensures that the tangent vector of the DLO at each subsequent edge can be described using the rotation matrix and the tangent vector at the previous edge.
Though we do not prove it here, the second condition ensures that the rotation between successive Bishop Frames is minimal while moving along the DLO.
In particular, this latter condition ensures that there is no twist about the subsequent tangent vectors at each edge of the DLO.
Finally, $\mathbf{b}_2^i := \mathbf{t}^i \times \mathbf{b}_1^i$, which ensures that  $\mathbf{b}_2^i$ is orthogonal to $\mathbf{t}^i$ and $\mathbf{b}_1^i$.

\begin{figure}[t]
    \centering
     \includegraphics[width=0.48\textwidth]{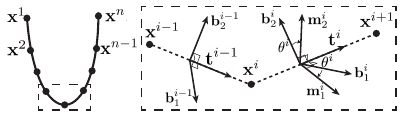}
     \caption{An illustration of DER coordinate frames.}
    \label{fig:DER_frames}
\end{figure}

\textbf{Material Frame:}
Subsequently, DER theory introduces Material Frames that are made up of three axes  $\{\mathbf{t}^i, \mathbf{m}_1^i, \mathbf{m}_2^i\}$.
These Material Frames are generated by rotating the Bishop frame about the tangent vector by a scalar $\theta^i$, i.e.,  $\mathbf{m}_1^i, \mathbf{m}_2^i$ are defined as
\begin{equation}
    \mathbf{m}_1^i =  \textbf{b}_1^i \cos \theta^i +  \textbf{b}_2^i \sin \theta^i
   \label{eq:m1}
\end{equation}
\begin{equation}
    \quad\mathbf{m}_2^i = -\textbf{b}_1^i \sin \theta^i +  \textbf{b}_2^i\cos \theta^i
   \label{eq:m2}
\end{equation}
An illustration of the frames can be found in Figure \ref{fig:DER_frames}.

\textbf{Equations of Motion:}
Let $\MaterialFrame(\mathbf{X}_t, \bm{\theta}_t)$ denote the set of all $n-1$ Material Frames at time $t$, where $\bm{\theta}_t = \{\theta_t^i\}_{i=2}^{n-1}$..
Let $\materialp$ represent the vector of material properties for each vertex of the DLO (i.e., mass, bending modulus, and twisting modulus) \citep[Section 4.2]{originalDER}.
Using these definitions, one can compute the potential energy of the DLO arising from bending and twisting which we denote by $P(\MaterialFrame(\mathbf{X}_t, \bm{\theta}_t), \materialp)$.
DER theory assumes that the DLO reaches an equilibrium state between each simulation time step to obtain the optimal $\bm{\theta}_t^*$:
\begin{equation} \bm{\theta}^*_t(\textbf{X}_t,\materialp) = \argmin_{\bm{\theta}_t} P(\MaterialFrame(\mathbf{X}_t, \bm{\theta}_t), \materialp)
       \label{eq:opt_theta}
\end{equation}
Once $\bm{\theta}^*_t(\mathbf{X}_t, \materialp)$ is derived, the restorative force during deformation is given by the negative gradient of the potential energy with respect to the vertices.
Let $\Massmatrix$ denote the mass matrix assembled from the per-vertex masses in $\materialp$.
Consequently, the equation of motion for the DLO is:
\begin{equation}
       \Massmatrix \ddot{\textbf{X}}_{t} = -\frac{\partial P(\MaterialFrame(\textbf{X}_t, \bm{\theta}^*_t(\textbf{X}_t,\materialp)),\bm{\alpha})}{\partial  \textbf{X}_{t}}
       \label{eq:equationofmotion}
\end{equation}
One can numerically integrate this formula to predict the velocity and position by applying the Semi-Implicit Euler method:
\begin{align}
        \hat{\textbf{V}}_{t+1} &= \hat{\textbf{V}}_\text{t}-\Delta_\text{t}\Massmatrix^{-1}\frac{\partial P(\MaterialFrame(\textbf{X}_t, \bm{\theta}^*_t(\textbf{X}_t,\materialp)),\bm{\alpha})}{\partial  \textbf{X}_{t}}, \label{eq:Semi-Euler1}\\
        \hat{\textbf{X}}_{\text{t}+1} &= \hat{\textbf{X}}_\text{t} + \Delta_\text{t}\hat{\textbf{V}}_{t+1},
       \label{eq:Semi-Euler2}
\end{align}
where $\Delta_{\text{t}} > 0$ is a user specified time discretization.

\subsubsection{Differentiable Discrete Elastic Rods (DDER)}
Tuning the DER model parameters, $\materialp$, to match real-world DLOs is non-trivial.
To address this challenge, DEFORM~\citep{DEFORM} introduces DDER, which enables the automatic tuning of DLO parameters using real-world data.
Predicting $\hat{\textbf{X}}_{t+1}$ with DDER through \eqref{eq:Semi-Euler1} and \eqref{eq:Semi-Euler2} involves solving an optimization problem as defined in \eqref{eq:opt_theta}.
Consequently, minimizing the prediction loss requires solving a bi-level optimization problem at each time step $t$:
\begin{align}
    \label{eq:outerloop}
    &&\underset{\bm{\alpha}}{\min} \hspace{0.5cm}& \|\textbf{X}_{t+1}- \hat{\textbf{X}}_{t+1}(\bm{\theta}^*_t)\|_2 \hspace{1.5cm}    \\
    && & \bm{\theta}^*_t = \argmin_{\bm{\theta}_t}   P(\MaterialFrame(\textbf{X}_t, \bm{\theta}_t),\materialp)
    \label{eq:innerloop}
\end{align}
Note $\hat{\textbf{X}}_{t+1}$ depends on the previously computed $\bm{\theta}^*_t$ through \eqref{eq:equationofmotion}, \eqref{eq:Semi-Euler1} and \eqref{eq:Semi-Euler2}.
To compute the gradient of \eqref{eq:innerloop} with respect to $\bm{\alpha}$, DDER employs solvers from Theseus \citep{theseus}, which use PyTorch’s automatic differentiation \citep{pytorch}.
As we illustrate via experiments in Section \ref{sec:Computational Speed}, this can be a time-consuming process.

\section{Methodology}

\subsection{Branched Deformable Linear Object (BDLO) Model}
\label{sec:BDLO_model}
\begin{figure}[t]
    \centering
    \includegraphics[width=0.4\textwidth]{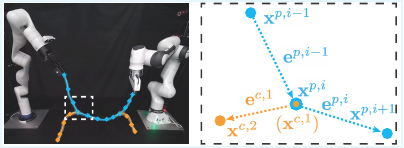}
    \caption{
    An illustration of a branched deformable linear object (BDLO) model.
    Left: An example configuration of BDLO manipulation.
    Right: Relevant notation and structure at a junction.}
    \label{fig:BDLO}
\end{figure}

This section defines the key concepts and notations used in BDLO modeling.
To model a BDLO, we decompose its structure into a hierarchical arrangement of DLOs.
The parent DLO is defined as the minimum-length path through the BDLO that connects the two end-effectors (i.e., the points where the structure is manipulated or anchored).
All other branches diverging from this primary path are modeled as children DLOs that maintain their physical connection to the parent DLO at their attachment points, which we refer to as \emph{junction points}.
The parent and children DLOs are each modeled as DLOs and inherit all previously introduced notation.
For conciseness, we describe the junction point using only three edges at each time instance: two from the parent DLO and one from a child DLO.
In principle, the methods in DEFT can be extended to handle junctions with more than three edges.

We distinguish between the parent and child DLOs using superscripts $p$ and $c$, respectively.
For instance, if a junction is located at the $i$-th vertex, the associated edges are denoted by $\timevertind{e}{p, i-1}{t}$, $\timevertind{e}{p, i}{t}$, and $\timevertind{e}{c, 1}{t}$.
An illustration of a BDLO can be found in Figure \ref{fig:BDLO}.
To capture bending–twisting interactions at junctions, we associate each edge $\timevertind{e}{i}{t}$ with an orientation described in world frame by an angle–axis vector $\mathbf{\Omega}^i_t \in \mathbb{R}^3$.
In the absence of accurate orientation tracking and prediction, a junction reduces to a purely translational link that transmits only stretching forces.
As demonstrated in our ablation study, this omission substantially degrades prediction accuracy for child branches and decreases the success rate of manipulation tasks.

\begin{figure}[t]
    \centering
\includegraphics[width=0.4\textwidth]{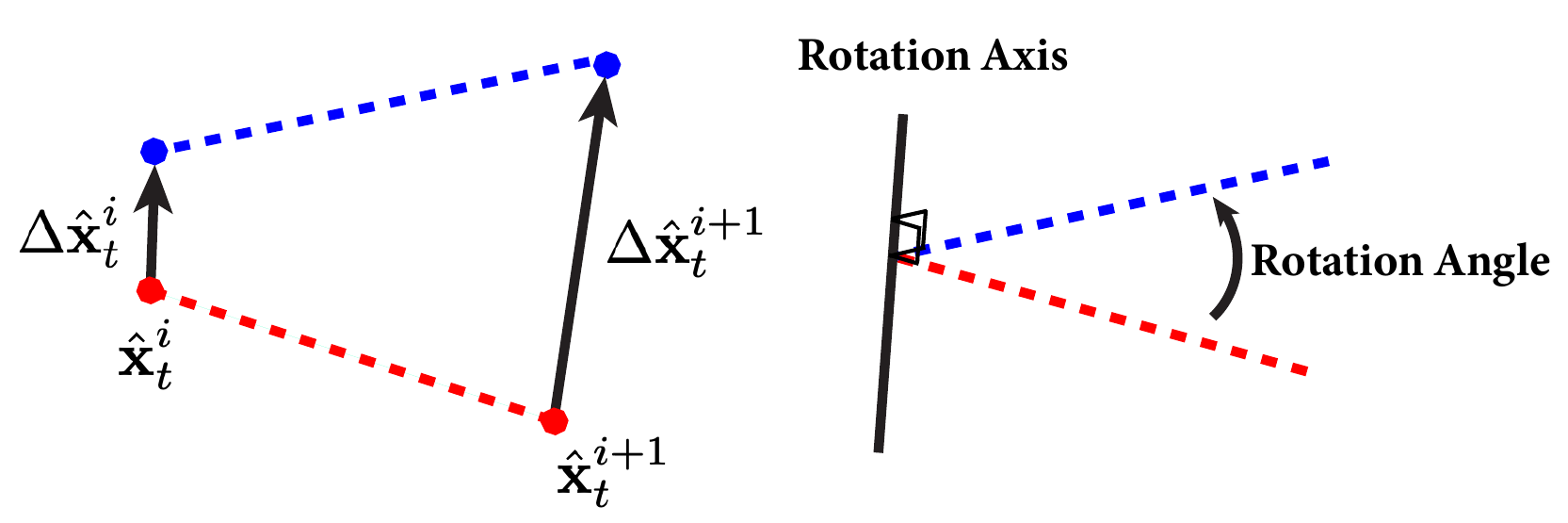}
    \caption{An illustration of $\hat{\boldsymbol{\Omega}}^i_{\Delta,t} (\hat{\mathbf{X}}_t,\Delta \hat{\mathbf{X}}_t)$
    }
    \label{fig:orientation change}
\end{figure}

However, in practice, it is both challenging and impractical to mount hardware on every segment to directly track orientation changes \citep{YANG2022104258}.
A key observation is that the orientation change $\hat{\boldsymbol{\Omega}}_{\Delta, t}$ arises from the displacement $\Delta \hat{\mathbf{X}}_t$ modifying the original configuration $\hat{\mathbf{X}}_t$.
Here, $\Delta \hat{\mathbf{X}}_t$ is induced by internal forces and the enforcement of constraints.
This insight allows us to approximate $\hat{\boldsymbol{\Omega}}_{\Delta, t}$ using only the known variables $\Delta \hat{\mathbf{X}}_{t}$ and $\hat{\mathbf{X}}_{t}$, as illustrated in Figure~\ref{fig:orientation change}.
Consequently, we define the incremental orientation change of segment $i$ as
\begin{align}
\hat{\boldsymbol{\Omega}}^i_{\Delta, t}(\hat{\mathbf{X}}_t, \Delta \hat{\mathbf{X}}_t)
&= \frac{\hat{\mathbf{x}}^{i+1}_t - \hat{\mathbf{x}}^i_t}
        {\|\hat{\mathbf{x}}^{i+1}_t - \hat{\mathbf{x}}^i_t\|_2} \times
\notag \\
&\hspace{0mm}
\frac{(\hat{\mathbf{x}}^{i+1}_t + \Delta\hat{\mathbf{x}}^{i+1}_t) - (\hat{\mathbf{x}}^i_t + \Delta\hat{\mathbf{x}}^i_t)}
     {\|(\hat{\mathbf{x}}^{i+1}_t + \Delta\hat{\mathbf{x}}^{i+1}_t) - (\hat{\mathbf{x}}^i_t + \Delta\hat{\mathbf{x}}^i_t)\|_2}.
\label{eq:delta_omega}
\end{align}
More precisely, for two unit tangent vectors separated by an angle \(\theta\), their cross product is directed along the rotation axis and has magnitude \(\sin\theta\). Equation~\eqref{eq:delta_omega} therefore uses the small-angle approximation \(\sin\theta\approx\theta\), whose absolute error satisfies \(0\leq\theta-\sin\theta\leq\theta^3/6\). This approximation is appropriate for the small per-step rotations produced using \(\Delta_t=0.01\)~s.

To simplify notation, we abbreviate $\hat{\boldsymbol{\Omega}}^i_{\Delta, t}(\hat{\mathbf{X}}_t, \Delta \hat{\mathbf{X}}_t) $ as $\hat{\boldsymbol{\Omega}}^i_{\Delta, t} $.
The orientation is updated on the Lie group $\mathrm{SO}(3)$ by
\begin{equation}
\label{eq:omega}
\hat{\boldsymbol{\Omega}}^{\,i}_{t+1}
= \Log\!\Big(
\Exp([\hat{\boldsymbol{\Omega}}^{\,i}_{\Delta,t}]_\times)\,
\Exp([\hat{\boldsymbol{\Omega}}^{\,i}_{t}]_\times)
\Big)^\vee ,
\end{equation}
where $\Exp$ and $\Log$ denote the exponential and logarithm maps between
$\mathfrak{so}(3)$ and $\mathrm{SO}(3)$ \citep{etingof2024liegroupsliealgebras}, and $(\cdot)^\vee$ extracts the vector from a skew-symmetric matrix.
The rotational dynamics of segment $i$ further depend on its moment of inertia $\moi \in \mathbb{R}^{3\times 3}$, which characterizes how the mass distribution resists angular deformation.
Let the orientation and position of the two end effectors at time $t$ be denoted by $\vertindu_t \in \R^{12}$.
We refer to $\vertindu_t$ as the input.

\subsection{DEFT Overview}
\begin{figure*}[t]
    \centering
    \includegraphics[width=1\textwidth]{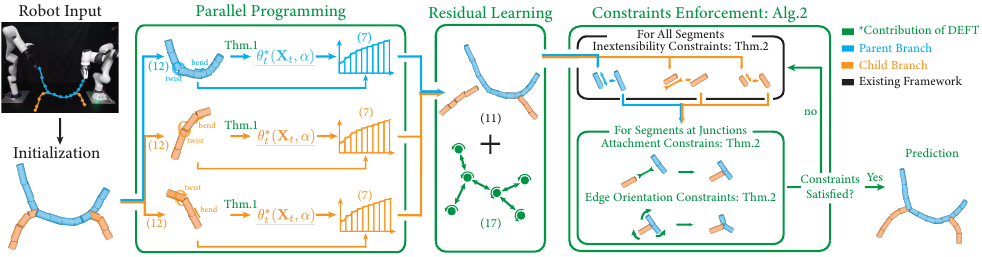}
    \caption{
    Algorithm Overview of DEFT.
    In the initialization stage (Section \ref{sec:BDLO_model}), DEFT begins by separating the BDLO into a parent DLO and one or more children DLOs.
    Each DLO is discretized into vertices and represented as elastic rods.
    This setup allows DEFT to capture the geometric and physical properties required for dynamic simulation.
    To improve computational efficiency, DEFT then predicts the dynamics of each branch in parallel (Section \ref{section:gradientopt}).
    During this process, analytical gradients are provided to solve \eqref{eq:innerloop}, as detailed in Section \ref{section:gradientopt}, ensuring efficient and stable convergence.
    Next, to address numerical errors, DEFT employs a GNN designed to learn the BDLO’s residual dynamics (Section \ref{section:residual learning}).
    By modeling discrepancies between simulated and observed behavior, the GNN refines predictions and enhances overall accuracy.
    After integration, DEFT enforces constraints (Section \ref{section:constratins}) to enforce physical realism.
    Inextensibility constraints are applied to each branch, while junction-level constraints ensure proper attachment at branch junctions.
    Additionally, edge orientation constraints enable the propagation of dynamics across these junctions.
    Throughout the entire pipeline, all components remain fully differentiable, allowing for efficient parameter learning from real-world data.
    }
    \label{fig:deft_contribution_diagram}
\end{figure*}

At each time $t$, DEFT simulates the behavior of a BDLO when given predictions of the vertices and their velocities under a specified input as described in Algorithm \ref{alg:long time horizon}.
It begins by first associating the grasped vertices with the robot end effector inputs  $\vertindu_t$.
DEFT then computes their respective material frames, which are subsequently used to calculate each branch’s potential energy (Line \ref{algo:line2}).
DEFT then minimizes this potential energy using analytical gradients to determine the optimal parameters $\bm{\theta}_t^*$ and updates the potential energy accordingly (Line \ref{algo:line3}).
These forces are integrated forward in time using a Semi-Implicit Euler method combined with residual learning to update the vertex velocities and positions (Lines \ref{algo:line4} and \ref{algo:line5}).
To further accelerate above computation, DEFT processes the parent and child DLOs in parallel.

However, due to numerical integration errors, branch lengths may change in length.
In addition, up to this point, DEFT treats each branch’s internal dynamics in isolation, without accounting for junction interactions.
This limits the propagation of dynamics and may even cause branch detachment.

To overcome this, DEFT enforces constraints ensuring branch inextensibility, junction connectivity, and proper propagation of deformation effects (Line \ref{algo:line6}).
Applying Algorithm \ref{alg:long time horizon} iteratively enables accurate BDLO state predictions over extended time horizons.
An algorithm overview is illustrated in Figure \ref{fig:deft_contribution_diagram}.
\begin{algorithm}[t]
\caption{$(\pred{X}{t+1}, \pred{V}{t+1}, \hat{\mathbf{\Omega}}_{t+1}) = \textbf{DEFT}(\pred{X}{t}, \pred{V}{t}, \hat{\mathbf{\Omega}}_t, \textbf{u}_t$)}
\label{alg:long time horizon}
\begin{algorithmic}[1]
\State \label{algo:line1} Associate grasped vertices with $\textbf{u}_t$
\State \label{algo:line2} Calculate Material Frames $\MaterialFrame(\textbf{X}_t, \bm{\theta}_t)$ \Comment{\eqref{eq:m1},\eqref{eq:m2}}
\State \label{algo:line3} $\bm{\theta}^*_t(\textbf{X}_t,\materialp) = \argmin_{\bm{\theta}_t} P(\MaterialFrame(\textbf{X}_t, \bm{\theta}_t),\materialp)$ \Comment{\secref{section:gradientopt}}
\State \label{algo:line4} $\hat{\textbf{V}}_{t+1} = \hat{\textbf{V}}_t-\Delta_t \Massmatrix^{-1}\frac{\partial P(\MaterialFrame(\textbf{X}_t, \bm{\theta}^*_t(\textbf{X}_t,\materialp)),\bm{\alpha})}{\partial \textbf{X}_{t}}$ \Comment{\eqref{eq:Semi-Euler1}}
\State \label{algo:line5} $\hat{\mathbf{X}}_{t+1} = \hat{\mathbf{X}}_t + \Delta_t\hat{\textbf{V}}_{t+1} + \text{GNN}$ \Comment{\secref{section:residual learning}} \label{alg1:residual}
\State \label{algo:line6} Constraints enforcement on $\hat{\mathbf{X}}_{t+1}$, $\hat{\mathbf{\Omega}}_t$ \Comment{Algo.\ref{alg:constraints enforcement}}
\State \label{algo:line7} $\hat{\textbf{V}}_{t+1} = (\hat{\mathbf{X}}_{t+1} - \hat{\mathbf{X}}_t)/\Delta_t$ \Comment{Velocity Update}
\end{algorithmic}
\Return $\pred{X}{t+1}, \pred{V}{t+1},  \hat{\mathbf{\Omega}}_{t+1}$
\end{algorithm}
\subsection{Integration Method with Residual Learning}
\label{section:residual learning}

\begin{figure}
    \centering
    \includegraphics[width=0.45\textwidth]{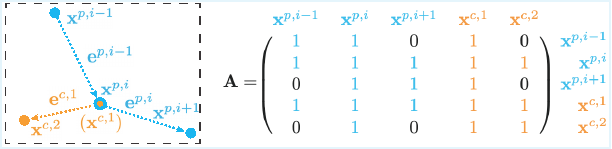}
    \caption{
    The adjacency matrix $\mathbf{A}$ for the junction in Figure~\ref{fig:BDLO}
    captures both self-loops (diagonal entries) and inter-node connections
    (off-diagonal entries). By embedding each node’s local dynamics
    alongside its coupling to neighboring nodes, $\mathbf{A}$ enables
    a graph representation of the BDLO’s behavior.}
    \label{fig:matrix_A_illustration}
\end{figure}

This subsection constructs the GNN for residual learning in integration method.
The motivation for employing a GNN lies in its ability to represent a BDLO as a graph, where each vertex is treated as a node and edges capture both intra-branch connections and junctions.
This graph-based formulation translates into an adjacency matrix \(\mathbf{A} \in \mathbb{R}^{(n^p + n^c) \times (n^p + n^c)}\), which links individual branch nodes and explicitly encodes node information across branch junctions.
This adjacency matrix is defined as:
\begin{equation}
\mathbf{A}_{i_1, i_2} =
\begin{cases}
1 & \text{if there is an edge between } i_1 \text{ and } i_2, \\
0 & \text{otherwise,}
\end{cases}
\end{equation}
An illustration of $\mathbf{A}$ can be found in Figure \ref{fig:matrix_A_illustration}.
In practice,  DEFT constructs $\mathbf{A}$ for each BDLO whose behavior a user wants to predict.
This adjacency matrix is utilized within a Graph Convolutional Network (GCN)~\citep{GCN} to aggregate learned features.
A single GCN layer is defined as
\begin{equation}
  \text{GCN}(\hat{\mathbf{X}}_t, \hat{\mathbf{V}}_t, \bm{\alpha})
  = \mathbf{\tilde{A}} \,\mathbf{F}_t(\hat{\mathbf{X}}_t, \hat{\mathbf{V}}_t, \bm{\alpha}) \,\mathbf{W},
  \label{eq:GCN}
\end{equation}
where \(\mathbf{\tilde{A}}\) is the normalized adjacency matrix
\begin{equation}
\mathbf{\tilde{A}} = \mathbf{D}^{-\frac{1}{2}} \mathbf{A} \mathbf{D}^{-\frac{1}{2}}.
\label{eq:normalizedA}
\end{equation}
Note that $\mathbf{\tilde{A}}$ is the normalization of \(\mathbf{A}\) using the degree matrix \(\mathbf{D} = \mathrm{diag}(d_1, d_2, \ldots, d_n)\)  where \(d_{i1} = \sum_{i2} A_{i1,i2}\).
This normalization ensures that nodes with a high degree do not disproportionately influence the overall performance.
This ensures consistent feature magnitudes across the learned representation~\citep{GNNreview}.
The node feature matrix, \(\mathbf{F}_t(\hat{\mathbf{X}}_t, \hat{\mathbf{V}}_t, \bm{\alpha}) \in \mathbb{R}^{(n^p + n^c) \times h}\), is constructed by concatenating its arguments together.
The learnable feature transformation, parameterized by \(\mathbf{W}\), is implemented as a multilayer perceptron (MLP)~\citep{MLP}, which maps input features to residual corrections. Its architecture and training configuration are summarized in Table~\ref{tab:mlp_training_details}.

\subsection{Enforcing Constraints}
\label{section:constratins}

After performing numerical integration, DEFT applies constraints to ensure branch inextensibility, ensure that branches are attached at the junctions, and ensure that deformations are propagated throughout the BDLO.
To accomplish this, DEFT updates the positions of each vertex using correction terms $\Delta \pred{X}{}$.
One could just update the positions of each vertex by projecting directly onto the constraint set.
However, this induces highly nonlinear changes in momentum and can lead to simulation instability and unrealistic behavior.

To address these challenges, DEFT enforces the aforementioned constraints while trying to preserve momentum.
DEFT does this by formulating the computation of the correction term as the solution to an optimization problem.
As we illustrate through our experimental evaluation in Section \ref{sec:experiments}, this leads to more stable simulations and efficient parameter learning, which corresponds to more realistic predictions.
Finally, note that DEFT could formulate the optimization problem to compute the correction terms as a single large optimization problem that considers constraints between all vertices in the BDLO simultaneously.
However, this large optimization problem can be difficult to solve rapidly.
Instead DEFT enforces these constraints sequentially between neighboring nodes.
Next, we summarize each of these optimization problems.

\subsubsection{Inextensibility Enforcement}

To enforce inextensibility of each branch, we define a constraint function between successive vertices.
This function penalizes deviations of each segment’s length from its undeformed length.
By applying the following theorem, one can compute $\Delta \hat{\mathbf{x}}^i, \Delta \hat{\mathbf{x}}^{i+1}$ explicitly:
\begin{thm}[Computing the Corrections to Enforce Inextensibility]
    \label{theorem:inextensibility}
    Consider two successive vertices \(i\) and \(i+1\) along the same branch.
    Let \(\overline{\mathbf{e}}_i\) denote the edge between these vertices when the BDLO is undeformed.
    Let the constraint function to enforce inextensibility be defined as:
    \begin{equation}
   \hspace*{-0.1cm} C^i_{I}(\hat{\mathbf{X}},\Delta \hat{\mathbf{X}} ) =
    ||(\hat{\mathbf{x}}^i+\Delta \hat{\mathbf{x}}^i) - (\hat{\mathbf{x}}^{i+1}+\Delta \hat{\mathbf{x}}^{i+1})||_2 -  ||\overline{\textbf{e}}_i||_2.
    \label{eq:constraint_inextensibility}
    \end{equation}
    The correction $\Delta \hat{\mathbf{X}}$ to enforce inextensibility at joint $i$ can be computed by solving the following optimization problem:
    \begin{align}
    &&\underset{\Delta \hat{\mathbf{X}}}{\min} & \hspace{0.3cm} \frac{1}{2}\left(C^i_{I}(\hat{\mathbf{X}},\Delta \hat{\mathbf{X}} ) \right)^2   & \label{eq:cost} \\
    &&\text{s.t.} &\hspace{0.3cm}  \Delta \hat{\mathbf{x}}^j = 0,   \hspace{0.3cm} \forall j \notin \{i,i+1\}, \notag  \\
    && & \hspace{0.3cm}  \mathbf{M}^i  \Delta \hat{\mathbf{x}}^i + \mathbf{M}^{i+1}  \Delta \hat{\mathbf{x}}^{i+1}= \mathbf{0}, &  \notag \\
    && &\hspace{0.3cm}
    \mathbf{I}^i   \hat{\boldsymbol{\Omega}}^i_\Delta (\hat{\mathbf{X}},\Delta \hat{\mathbf{X}} ) = \mathbf{0}, & \notag
\end{align}
where $\hat{\boldsymbol{\Omega}}^i_\Delta (\hat{\mathbf{X}},\Delta \hat{\mathbf{X}} ) \in \mathbb{R}^{3}$ is defined in \ref{eq:delta_omega} and represents the orientation change of the $i$th edge arising from $\Delta \hat{\mathbf{x}}^i$ and $\Delta \hat{\mathbf{x}}^{i+1}$ and $\mathbf{0}$ correspond to the zero vector of appropriate size, and $\mathbf{M}^i$ is the per-vertex mass matrix of vertex $i$, assembled from the per-vertex masses in $\bm{\alpha}$ (consistent with the global mass matrix $\mathbf{M}$ in \eqref{eq:equationofmotion}).
The only non-zero elements of the minimizer to this optimization problem can be computed explicitly and are equal to:
\begin{equation}
    \label{eq:momentum_solution1_inext}
    \Delta \hat{\mathbf{x}}^i
    =
    \mathbf{M}^{i+1}\bigl(\mathbf{M}^i + \mathbf{M}^{i+1}\bigr)^{-1}
    C^i_{I}(\hat{\mathbf{X}}, \mathbf{0})
    \frac{ \bigl(\hat{\mathbf{x}}^{i+1} - \hat{\mathbf{x}}^{i}\bigr)
         }{\|\hat{\mathbf{x}}^{i+1} - \hat{\mathbf{x}}^{i}\|_2},
\end{equation}
\begin{equation}
    \label{eq:momentum_solution2_inext}
    \Delta \hat{\mathbf{x}}^{i+1}
    =
    \mathbf{M}^i(\mathbf{M}^i + \mathbf{M}^{i+1}\bigr)^{-1}
       C^i_{I}(\hat{\mathbf{X}}, \mathbf{0})
    \frac{ \bigl(\hat{\mathbf{x}}^{i} - \hat{\mathbf{x}}^{i+1}\bigr)
         }{\|\hat{\mathbf{x}}^{i+1} - \hat{\mathbf{x}}^{i}\|_2}.
\end{equation}
\end{thm}
The inextensibility constraint at vertex $i$ is satisfied when $C^i_{I}(\hat{\mathbf{X}}, \mathbf{0})^2 = 0$.
For convenience, let the function that computes the minimizer of optimization problem \eqref{eq:cost} using $\hat{\mathbf{X}}$ be denoted by $D^i_I$ (i.e., $D^i_I: \hat{\mathbf{X}} \mapsto \Delta \hat{\mathbf{X}}$).

Please note that Theorem~\ref{theorem:inextensibility} provides an alternative
formulation to \citep[Section~4.3]{DEFORM}.
The main difference lies in the
zero-angular-displacement condition,
\(
\mathbf I^i\hat{\boldsymbol{\Omega}}_\Delta^i
(\hat{\mathbf X},\Delta\hat{\mathbf X})
=
\mathbf 0
\).
This alternative form enables the enforcement of all of the following constraints within a unified framework. Theorem~\ref{theorem:inextensibility} can still be validated by following the
same proof strategy as in \citep[Appendix~B.2]{DEFORM}.

\subsubsection{Enforcing Attachment at the Junction}
\label{sec:Attachment Enforcement at the Junction}
To ensure the child branch remains attached to the parent branch, we define a constraint function between their junction vertices.
By applying the following theorem, one can compute $\Delta \hat{\mathbf{x}}^{p,i}, \Delta \hat{\mathbf{x}}^{c,1}$ explicitly:
\begin{thm}
\label{theorem:constraint_attachement}
Consider the parent-branch vertex $\hat{\mathbf{x}}^{p,i}$ and child-branch vertex $\hat{\mathbf{x}}^{c,1}$ at the junction such as the one depicted in Figure \ref{fig:BDLO}.
\label{thm:attachment}
Let the constraint function to enforce attachment at the junction be defined as:
\begin{equation}
\label{eq:constraint_attachement}
 C^{p,i}_A(\hat{\mathbf{X}}, \Delta \hat{\mathbf{X}}) = ||\bigl(\hat{\mathbf{x}}^{p,i} + \Delta \hat{\mathbf{x}}^{p,i}\bigr) - \bigl(\hat{\mathbf{x}}^{c,1} + \Delta \hat{\mathbf{x}}^{c,1}\bigr)||_2.
    \end{equation}
Suppose that the correction $\Delta \hat{\mathbf{X}}$ to account for the junction attachment constraint is obtained by solving an optimization problem that is identical to \eqref{eq:cost} with the objective function replaced with \eqref{eq:constraint_attachement} and with a requirement that the only elements of $\Delta \hat{\mathbf{X}}$ are those associated vertices $p,i$ and $c,1$.
The only non-zero elements of the minimizer to this optimization problem can be computed explicitly and are equal to:
\begin{equation*}
    \label{eq:momentum_solution1_attach}
  \Delta \hat{\mathbf{x}}^{p, i} =  \mathbf{M}^{c, 1} (\mathbf{M}^{p, i}+\mathbf{M}^{c, 1})^{-1}  C^{p,i}_A(\hat{\mathbf{X}},\mathbf{0}) \frac{(\hat{\mathbf{x}}^{c, 1} - \hat{\mathbf{x}}^{p, i})}{||\hat{\mathbf{x}}^{c, 1} - \hat{\mathbf{x}}^{p, i}||_2},
\end{equation*}
\begin{equation*}
    \label{eq:momentum_solution2_attach}
     \Delta \hat{\mathbf{x}}^{c, 1} =  \mathbf{M}^{p, i}  (\mathbf{M}^{p, i}+\mathbf{M}^{c, 1})^{-1}
    \ C^{p,i}_A(\hat{\mathbf{X}},\mathbf{0}) \frac{(\hat{\mathbf{x}}^{p, i} - \hat{\mathbf{x}}^{c, 1})}{||\hat{\mathbf{x}}^{c, 1} - \hat{\mathbf{x}}^{p, i}||_2}.
\end{equation*}
\end{thm}
Note that the junction attachment constraint is satisfied at vertex $p,i$ when $C^{p,i}_{I}(\hat{\mathbf{X}}, \mathbf{0})^2 = 0$.
For convenience, let the function that computes the minimizer of optimization problem \eqref{eq:cost} using $\hat{\mathbf{X}}$ be denoted by $D^{p,i}_A$ (i.e., $D^{p,i}_A: \hat{\mathbf{X}} \mapsto \Delta \hat{\mathbf{X}}$).

\textit{Proof of Theorem \ref{thm:attachment}.}
\label{appendix:theorem_junction1}
 We first formulate the optimization problem:
\begin{align}
    & &\underset{\Delta \hat{\mathbf{X}}}{\min}& \hspace{0.3cm} \frac{1}{2}\left(C^{p,i}_A(\hat{\mathbf{X}}, \Delta \hat{\mathbf{X}})\right)^2
    \label{eq:optimization_simplified_attachement}\\
    &&\text{s.t.} &\hspace{0.3cm}  \Delta \hat{\mathbf{x}}^j = \mathbf{0}, \hspace{0.3cm} \forall j \notin \{(p, i),(c, 1)\}
    \label{eq:index_selection}\\
    && & \hspace{0.3cm}   \mathbf{M}^{p, i}  \Delta \hat{\mathbf{x}}^{p, i}\ + \mathbf{M}^{c, 1}  \Delta \hat{\mathbf{x}}^{c, 1}= \mathbf{0},
    \label{eq:linearm_simplified} \\
    && &\hspace{0.3cm}
    \textbf{I}^{p, i}   \hat{\boldsymbol{\Omega}}^{p, i} _\Delta (\hat{\mathbf{X}},\Delta \hat{\mathbf{X}} ) = \mathbf{0},
    \label{eq:angularm_simplified}
\end{align}
where $\hat{\boldsymbol{\Omega}}^{p, i} _\Delta (\hat{\mathbf{X}},\Delta \hat{\mathbf{X}} )$ represents the orientation change of the edge arising from $\Delta \hat{\mathbf{x}}^{p, i}$ and $\Delta \hat{\mathbf{x}}^{c, 1}$.
To solve the above optimization, we introduce Lagrange multipliers $\boldsymbol{\lambda}_\text{l} \in \mathbb{R}^{3}$ and $\boldsymbol{\lambda}_\text{r} \in \mathbb{R}^{3}$ associated with constraints \eqref{eq:linearm_simplified} and \eqref{eq:angularm_simplified}, respectively.
The corresponding Lagrangian $\mathcal{L}$  can be found as follow:
\begin{align}
  & \mathcal{L}(\Delta \hat{\textbf{x}}^{p, i}, \Delta \hat{\mathbf{x}}^{c,1}, \boldsymbol{\lambda}_\text{l}, \boldsymbol{\lambda}_\text{r}) = \frac{1}{2}\left(C^{p,i}_A(\hat{\mathbf{X}}, \Delta \hat{\mathbf{X}})\right)^2 - \notag  \label{eq:attachment_lag} \\
  \hspace{-5mm} &   \boldsymbol{\lambda}^T_\text{l}(\Massmatrix^i \Delta \hat{\textbf{x}}^{p, i} + \Massmatrix^{c, 1} \Delta \hat{\mathbf{x}}^{c,1}) - \boldsymbol{\lambda}^T_\text{r}\textbf{I}^{p, i} \hat{\boldsymbol{\Omega}}^{p, i}_\Delta (\hat{\mathbf{X}}, \Delta \hat{\mathbf{X}}).
\end{align}
Next, we take the partial derivatives of \eqref{eq:attachment_lag} with respect to $\Delta \hat{\textbf{x}}^{p, i}, \Delta \hat{\mathbf{x}}^{c,1},
\boldsymbol{\lambda}_\text{l}, \boldsymbol{\lambda}_\text{r}$.
We define the unit displacement vector between the parent and child keypoints after
displacement as:
\begin{equation}
    \label{eq:unit_disp}
    \hat{\mathbf{d}}^{p,i} = \frac{(\hat{\mathbf{x}}^{p,i} + \Delta\hat{\mathbf{x}}^{p,i}) - (\hat{\mathbf{x}}^{c,1} + \Delta\hat{\mathbf{x}}^{c,1})}{\|(\hat{\mathbf{x}}^{p,i} + \Delta\hat{\mathbf{x}}^{p,i}) - (\hat{\mathbf{x}}^{c,1} + \Delta\hat{\mathbf{x}}^{c,1})\|_2}.
\end{equation}
Taking the partial derivatives of \eqref{eq:attachment_lag} with respect to
$\Delta \hat{\mathbf{x}}^{p,i}$, $\Delta \hat{\mathbf{x}}^{c,1}$, $\boldsymbol{\lambda}_\text{l}$,
and $\boldsymbol{\lambda}_\text{r}$ and setting each to zero yields:
\begin{align}
    \label{eq:lagarange1}
    \frac{\partial \mathcal{L}}
         {\partial \Delta\hat{\mathbf{x}}^{p,i}}
    &=
    C^{p,i}_A(\hat{\mathbf{X}},\Delta\hat{\mathbf{X}})
    \,\hat{\mathbf{d}}^{p,i}
    -
    \boldsymbol{\lambda}^{T}_{\text{l}}
    \mathbf{M}^{p,i}
    -
    \nonumber\\
    &\hspace{2.5cm}
    \frac{
        \partial
        \boldsymbol{\lambda}^{T}_{\text{r}}
        \mathbf{I}^{p,i}
        \hat{\boldsymbol{\Omega}}^{p,i}_{\Delta}
        (\hat{\mathbf{X}},\Delta\hat{\mathbf{X}})
    }{
        \partial\Delta\hat{\mathbf{x}}^{p,i}
    }
    =
    \mathbf{0}
    \\[6pt]
    \label{eq:lagarange2}
    \frac{\partial \mathcal{L}}
         {\partial \Delta\hat{\mathbf{x}}^{c,1}}
    &=
    -
    C^{p,i}_A(\hat{\mathbf{X}},\Delta\hat{\mathbf{X}})
    \,\hat{\mathbf{d}}^{p,i}
    -
    \boldsymbol{\lambda}^{T}_{\text{l}}
    \mathbf{M}^{c,1}
    -
    \nonumber\\
    &\hspace{2.5cm}
    \frac{
        \partial
        \boldsymbol{\lambda}^{T}_{\text{r}}
        \mathbf{I}^{p,i}
        \hat{\boldsymbol{\Omega}}^{p,i}_{\Delta}
        (\hat{\mathbf{X}},\Delta\hat{\mathbf{X}})
    }{
        \partial\Delta\hat{\mathbf{x}}^{c,1}
    }
    =
    \mathbf{0}
    \\[6pt]
    \label{eq:lagarangem}
    \frac{\partial \mathcal{L}}
         {\partial\boldsymbol{\lambda}_{\text{l}}}
    &=
    -
    \left(
        \mathbf{M}^{p,i}\Delta\hat{\mathbf{x}}^{p,i}
        +
        \mathbf{M}^{c,1}\Delta\hat{\mathbf{x}}^{c,1}
    \right)
    =
    \mathbf{0}
    \\[6pt]
    \label{eq:lagarangel}
    \frac{\partial \mathcal{L}}
         {\partial\boldsymbol{\lambda}_{\text{r}}}
    &=
    -\mathbf{I}^{p,i}
        \hat{\boldsymbol{\Omega}}^{p,i}_{\Delta}
        (\hat{\mathbf{X}},\Delta\hat{\mathbf{X}})
    =
    \mathbf{0}
\end{align}
We first observe that solving \eqref{eq:lagarangem} leads to the following equation:
\begin{equation}
     \Delta \hat{\textbf{x}}^{c, 1} = -(\Massmatrix^{c, 1})^{-1}\Massmatrix^{p, i}  \Delta \hat{\mathbf{x}}^{p,i}
     \label{eq:subsitituion}
\end{equation}
The explicit solutions above show that
\(\Delta\hat{\mathbf{x}}^{p,i}\) and
\(\Delta\hat{\mathbf{x}}^{c,1}\) are both collinear with the
parent--child edge direction.
Consequently, the corrected parent--child edge remains collinear
with its original direction.
Therefore, the cross product in \eqref{eq:delta_omega} vanishes,
yielding
\(\hat{\boldsymbol{\Omega}}_{\Delta}^{p,i}=\mathbf{0}\),
regardless of the correction magnitude.
Hence, the zero-angular-displacement condition is satisfied.
Next, summing \eqref{eq:lagarange1} with \eqref{eq:lagarange2} yields the following:
\begin{equation}
\boldsymbol{\lambda}_\text{l}\Massmatrix^{p, i} + \boldsymbol{\lambda}_\text{l}\Massmatrix^{c, 1} = \textbf{0}
\label{eq:lambdar0}
\end{equation}
As mass matrices are always positive definite.
This indicates $\boldsymbol{\lambda_l} = \textbf{0}$.
Based on \eqref{eq:lagarange1}, \eqref{eq:subsitituion} and \eqref{eq:lambdar0}, solving for $\Delta  \hat{\mathbf{x}}^{p,i}$ is equivalent to solving the following equation:
\begin{equation}
    \label{eq:constraint_derivation}
    C^{p,i}_A(\hat{\mathbf{X}}, \Delta \hat{\mathbf{X}}) = \textbf{0}
\end{equation}
By combining \eqref{eq:subsitituion} and \eqref{eq:constraint_derivation} and applying algebraic manipulations, we obtain:
\begin{equation*}
    \label{eq:momentum_solution1}
  \Delta \hat{\mathbf{x}}^{p, i} =  \mathbf{M}^{c, 1} (\mathbf{M}^{p, i}+\mathbf{M}^{c, 1})^{-1}  C^{p,i}_A(\hat{\mathbf{X}},\mathbf{0}) \frac{(\hat{\mathbf{x}}^{c, 1} - \hat{\mathbf{x}}^{p, i})}{||\hat{\mathbf{x}}^{c, 1} - \hat{\mathbf{x}}^{p, i}||_2},
\end{equation*}
Using $\Delta \hat{\mathbf{x}}^{p, i}$ and \eqref{eq:subsitituion}, one can find:
\begin{equation*}
    \label{eq:momentum_solution2}
     \Delta \hat{\mathbf{x}}^{c, 1} =  \mathbf{M}^{p, i}  (\mathbf{M}^{p, i}+\mathbf{M}^{c, 1})^{-1}
    \ C^{p,i}_A(\hat{\mathbf{X}},\mathbf{0}) \frac{(\hat{\mathbf{x}}^{p, i} - \hat{\mathbf{x}}^{c, 1})}{||\hat{\mathbf{x}}^{c, 1} - \hat{\mathbf{x}}^{p, i}||_2}.
\end{equation*}

\subsubsection{Edge Orientation Constraints at the Junction}

Many practical wire harnesses feature junctions that are nearly rigid. This near-rigidity arises because manufacturers often use rigid plastics or other stiff materials at these connection points \citep{wireharnesinstruction}. This results in minimal rotational or bending freedom. Consequently, the dynamics between the parent and child branches propagate through the junction in an almost rigid manner. To capture this behavior, we model each edge as a rigid body and define a constraint function that computes the relative orientation of each body. By applying the following theorem, one can compute \(\Delta \hat{\mathbf{x}}^{p,i}\) and \(\Delta \hat{\mathbf{x}}^{c,1}\) explicitly:

\begin{thm}
\label{thm:junctionconstraint}

Consider the parent-branch vertex \(\hat{\mathbf{x}}^{p,i}\) and child-branch vertex \(\hat{\mathbf{x}}^{c,1}\) at the junction shown in Figure~\ref{fig:BDLO}.

Let \(\epsilon\in[0,1]\) be a dimensionless relative-slack parameter.
Given the parent and child orientation corrections, their updated orientations are
\begin{equation}
\label{eq:constraints_orientation1}
\begin{aligned}
\hat{\bm\Omega}^{p,i}_{t+1}
&=
\Log\!\Big(
\Exp([\hat{\bm\Omega}^{p,i}_{\Delta}]_\times)
\Exp(
[\hat{\bm\Omega}^{p,i}_{t}]_\times
)
\Big)^\vee,
\\[4pt]
\hat{\bm\Omega}^{c,1}_{t+1}
&=
\Log\!\Big(
\Exp(
[\hat{\bm\Omega}^{c,1}_{\Delta}]_\times
)
\Exp(
[\hat{\bm\Omega}^{c,1}_{t}]_\times
)
\Big)^\vee.
\end{aligned}
\end{equation}

Define the accumulated parent and child rotations as
\begin{equation}
\label{eq:junction_rotation_matrices}
R_t^p
=
\Exp(
[\hat{\boldsymbol{\Omega}}^{p,i}_t]_\times
),
\qquad
R_t^c
=
\Exp(
[\hat{\boldsymbol{\Omega}}^{c,1}_t]_\times
),
\end{equation}
and their rotational mismatch as
\begin{equation}
\label{appendix:angular_difference}
\mathbf g\!\left(
\hat{\boldsymbol{\Omega}}^{p,i}_t,
\hat{\boldsymbol{\Omega}}^{c,1}_t
\right)
=
\Log\!\left(
R_t^c(R_t^p)^\top
\right)^\vee.
\end{equation}
The corresponding quantities at \(t+1\) are defined analogously.

The relative-slack orientation constraint is
\begin{equation}
\label{eq:constraints_orientation2}
C^{p,i}_O(\hat{\mathbf{X}},\Delta\hat{\mathbf{X}})
=
\left\|
\mathbf g\!\left(
\hat{\bm\Omega}^{p,i}_{t+1},
\hat{\bm\Omega}^{c,1}_{t+1}
\right)
-
\epsilon\,
\mathbf g\!\left(
\hat{\bm\Omega}^{p,i}_{t},
\hat{\bm\Omega}^{c,1}_{t}
\right)
\right\|_2.
\end{equation}

Here, \(\mathbf g\) measures the mismatch between rotations relative to the corresponding reference orientations, rather than the physical opening angle. The parameter \(\epsilon\) specifies the fraction of this mismatch retained after constraint enforcement; smaller values therefore produce a more rigid junction.

The orientation correction is obtained by solving
\begin{align}
&&
\underset{\Delta\hat{\mathbf{X}}}{\min}
&\quad
\frac{1}{2}
\left(
C^{p,i}_O(\hat{\mathbf{X}},\Delta\hat{\mathbf{X}})
\right)^2
&
\label{eq:constraint_rigid_body}
\\
&&
\text{s.t.}
&\quad
\mathbf I^{p,i}
\hat{\boldsymbol{\Omega}}_{\Delta}^{p,i}
+
\mathbf I^{c,1}
\hat{\boldsymbol{\Omega}}_{\Delta}^{c,1}
=
\mathbf 0.
&
\label{eq:constraint_rigid_body2}
\end{align}

Under the small-rotation approximation, explicit expressions for \(\hat{\boldsymbol{\Omega}}_{\Delta}^{p,i}\) and \(\hat{\boldsymbol{\Omega}}_{\Delta}^{c,1}\) can be derived and used to compute \(\Delta\hat{\mathbf{X}}\) through Rodrigues' rotation formula.~\citep{Rodrigues}

\end{thm}

Note that the junction orientation constraint is satisfied at vertex \(p,i\) when \(C^{p,i}_{O}(\hat{\mathbf{X}},\mathbf 0)^2=0\). For convenience, let the function that computes the minimizer of \eqref{eq:constraint_rigid_body} and \eqref{eq:constraint_rigid_body2} using \(\hat{\mathbf{X}}\) and \(\hat{\boldsymbol{\Omega}}\) be denoted by \(D^{p,i}_O\), i.e., \(D^{p,i}_O:\hat{\mathbf{X}}\mapsto\Delta\hat{\mathbf{X}}\).

\textit{Proof of Theorem \ref{thm:junctionconstraint}.}
\label{appendix:theorem_junction2}

\begin{figure}[t]
\centering
\includegraphics[width=0.3\textwidth]{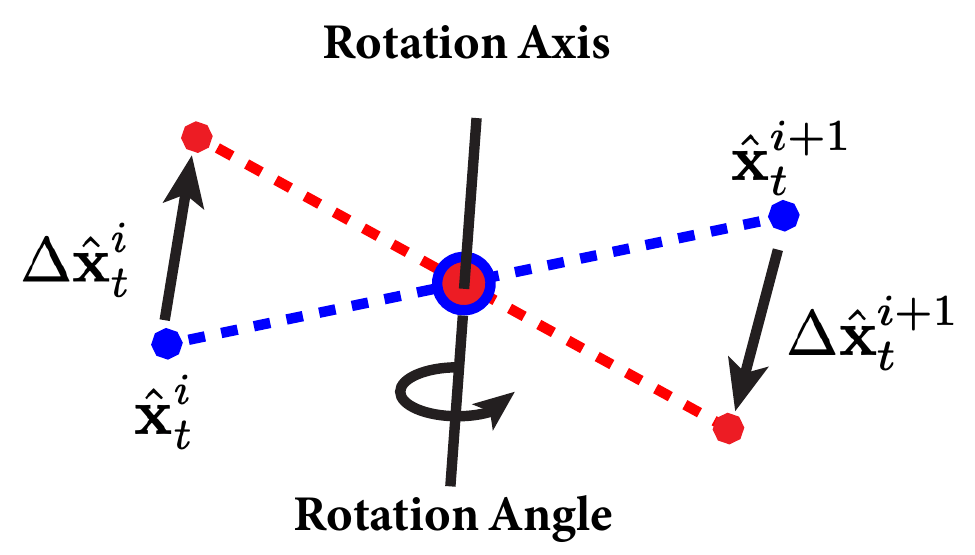}
\caption{An illustration of applying \(\hat{\boldsymbol{\Omega}}_{\Delta}^i\) to rotate the \(i\)-th segment.}
\label{fig:orientationonx}
\end{figure}

For convenience in the proof, we suppress the arguments to functions. To prove Theorem~\ref{thm:junctionconstraint}, we begin by describing how \(\hat{\boldsymbol{\Omega}}_{\Delta}^{i}\) is applied to rotate the \(i\)-th segment. Note that \(\Delta_t\) should be sufficiently small for \(\hat{\boldsymbol{\Omega}}_{\Delta}^{i}\) to accurately approximate the change in orientation. As shown in Figure~\ref{fig:orientationonx}, the rotation axis coincides with the center of the \(i\)-th segment. In practice, the rotation of the \(i\)-th segment is realized by converting the increment vector \(\hat{\boldsymbol{\Omega}}_{\Delta}^{i}\) into the rotation matrix \(\mathbf R_{\Delta}^{i}\) via Rodrigues' formula:
\begin{equation}
\label{eq:rodriguez}
\mathbf R_{\Delta}^{i}
=
\mathbf I
+
\frac{
\sin\|\hat{\boldsymbol{\Omega}}_{\Delta}^{i}\|
}{
\|\hat{\boldsymbol{\Omega}}_{\Delta}^{i}\|
}
[\hat{\boldsymbol{\Omega}}_{\Delta}^{i}]_\times
+
\frac{
1-\cos\|\hat{\boldsymbol{\Omega}}_{\Delta}^{i}\|
}{
\|\hat{\boldsymbol{\Omega}}_{\Delta}^{i}\|^2
}
[\hat{\boldsymbol{\Omega}}_{\Delta}^{i}]_\times^2,
\end{equation}
where \([\cdot]_\times\) denotes the skew-symmetric matrix associated with the cross product.

Consequently, the displacements of the adjacent vertices are obtained as
\begin{equation}
\Delta\hat{\mathbf{x}}^i
=
\mathbf R_{\Delta}^{i}
\frac{
\hat{\mathbf{x}}^{i+1}-\hat{\mathbf{x}}^i
}{2},
\label{eq:rx1}
\end{equation}
\begin{equation}
\Delta\hat{\mathbf{x}}^{i+1}
=
\mathbf R_{\Delta}^{i}
\frac{
\hat{\mathbf{x}}^i-\hat{\mathbf{x}}^{i+1}
}{2},
\label{eq:rx2}
\end{equation}

Because each segment rotates only about its own center, linear momentum is naturally conserved. Thus, we simplify the optimization problem analogously to \eqref{eq:constraint_attachement}, \eqref{eq:optimization_simplified_attachement}, and \eqref{eq:linearm_simplified}, resulting in \eqref{eq:constraints_orientation2}, \eqref{eq:constraint_rigid_body}, and \eqref{eq:constraint_rigid_body2}, respectively. By structural analogy, the proof mirrors that of Theorem~\ref{thm:attachment}, whereby solving \eqref{eq:constraint_rigid_body} and \eqref{eq:constraint_rigid_body2} reduces to enforcing two conditions:
\begin{equation}
\label{appendix:rotation_constraints}
C^{p,i}_O(
\hat{\mathbf{X}},
\Delta\hat{\mathbf{X}}
)
=
0,
\end{equation}
\begin{equation}
\label{appendix:angular_momentum_conserve}
\mathbf I^{p,i}
\hat{\boldsymbol{\Omega}}_{\Delta}^{p,i}
+
\mathbf I^{c,1}
\hat{\boldsymbol{\Omega}}_{\Delta}^{c,1}
=
\mathbf 0.
\end{equation}

Constraint~\eqref{appendix:rotation_constraints} requires the updated mismatch between the accumulated parent and child rotations to retain a fraction \(\epsilon\) of its current value.
Constraint~\eqref{appendix:angular_momentum_conserve} enforces angular momentum conservation. The task is therefore to distribute the correction in a way consistent with inertia balance.
Solving \eqref{appendix:angular_momentum_conserve} for the child correction gives:
\begin{equation}
\label{appendix:angular_momentum_conserve2}
\hat{\boldsymbol{\Omega}}_{\Delta}^{c,1}
=
-(\mathbf I^{c,1})^{-1}
\mathbf I^{p,i}
\hat{\boldsymbol{\Omega}}_{\Delta}^{p,i}.
\end{equation}
Using \eqref{appendix:angular_momentum_conserve2} together with the relative-slack constraint gives the following derivation.

The update rule in \eqref{eq:constraints_orientation1} premultiplies each current rotation by its corresponding correction rotation.
Enforcing the relative-slack constraint in \eqref{appendix:rotation_constraints} and applying the first-order Baker--Campbell--Hausdorff approximation under the small-rotation assumption gives
\begin{equation}
\label{appendix:relative_orientation_bch}
\begin{aligned}
\epsilon\,
\mathbf g\!\left(
\hat{\boldsymbol{\Omega}}^{p,i}_{t},
\hat{\boldsymbol{\Omega}}^{c,1}_{t}
\right)
&=
\mathbf g\!\left(
\hat{\boldsymbol{\Omega}}^{p,i}_{t+1},
\hat{\boldsymbol{\Omega}}^{c,1}_{t+1}
\right)
\\
&=
\Log\!\Big(
\Exp([\hat{\boldsymbol{\Omega}}_{\Delta}^{c,1}]_\times)
R_t^c(R_t^p)^\top
\\
&\qquad
\Exp(-[\hat{\boldsymbol{\Omega}}_{\Delta}^{p,i}]_\times)
\Big)^\vee
\\
&\approx
\mathbf g\!\left(
\hat{\boldsymbol{\Omega}}^{p,i}_{t},
\hat{\boldsymbol{\Omega}}^{c,1}_{t}
\right)
+
\hat{\boldsymbol{\Omega}}_{\Delta}^{c,1}
-
\hat{\boldsymbol{\Omega}}_{\Delta}^{p,i}.
\end{aligned}
\end{equation}
Rearranging \eqref{appendix:relative_orientation_bch} gives
\begin{equation}
\label{appendix:relative_orientation_correction}
\hat{\boldsymbol{\Omega}}_{\Delta}^{p,i}
-
\hat{\boldsymbol{\Omega}}_{\Delta}^{c,1}
=
(1-\epsilon)
\mathbf g\!\left(
\hat{\boldsymbol{\Omega}}^{p,i}_{t},
\hat{\boldsymbol{\Omega}}^{c,1}_{t}
\right).
\end{equation}
Substituting \eqref{appendix:angular_momentum_conserve2} into \eqref{appendix:relative_orientation_correction} gives
\begin{equation}
\label{appendix:parent_orientation_linear_system}
\left(
\mathbf I
+
(\mathbf I^{c,1})^{-1}\mathbf I^{p,i}
\right)
\hat{\boldsymbol{\Omega}}_{\Delta}^{p,i}
=
(1-\epsilon)
\mathbf g\!\left(
\hat{\boldsymbol{\Omega}}^{p,i}_{t},
\hat{\boldsymbol{\Omega}}^{c,1}_{t}
\right).
\end{equation}
Multiplying both sides of \eqref{appendix:parent_orientation_linear_system} by \(\mathbf I^{c,1}\) gives
\begin{equation}
\label{appendix:parent_orientation_weighted_system}
(\mathbf I^{p,i}+\mathbf I^{c,1})
\hat{\boldsymbol{\Omega}}_{\Delta}^{p,i}
=
(1-\epsilon)
\mathbf I^{c,1}
\mathbf g\!\left(
\hat{\boldsymbol{\Omega}}^{p,i}_{t},
\hat{\boldsymbol{\Omega}}^{c,1}_{t}
\right).
\end{equation}
Because the moment-of-inertia matrices used in our formulation are diagonal, they commute with their sum and its inverse. Therefore, solving \eqref{appendix:parent_orientation_weighted_system} for \(\hat{\boldsymbol{\Omega}}_{\Delta}^{p,i}\) gives

\begin{equation}
\label{eq:momentum_solution1_ore}
\hat{\boldsymbol{\Omega}}_{\Delta}^{p,i}
=
(1-\epsilon)
\mathbf I^{c,1}
(\mathbf I^{p,i}+\mathbf I^{c,1})^{-1}
\mathbf g(
\hat{\boldsymbol{\Omega}}^{p,i}_t,
\hat{\boldsymbol{\Omega}}^{c,1}_t
).
\end{equation}

Rearranging \eqref{appendix:relative_orientation_correction} gives
\begin{equation}
\label{appendix:child_orientation_from_parent}
\hat{\boldsymbol{\Omega}}_{\Delta}^{c,1}
=
\hat{\boldsymbol{\Omega}}_{\Delta}^{p,i}
-
(1-\epsilon)
\mathbf g\!\left(
\hat{\boldsymbol{\Omega}}^{p,i}_{t},
\hat{\boldsymbol{\Omega}}^{c,1}_{t}
\right).
\end{equation}
Substituting \eqref{eq:momentum_solution1_ore} into \eqref{appendix:child_orientation_from_parent} and using
\begin{equation}
\label{appendix:child_inertia_identity}
\mathbf I^{c,1}
(\mathbf I^{p,i}+\mathbf I^{c,1})^{-1}
-
\mathbf I
=
-\mathbf I^{p,i}
(\mathbf I^{p,i}+\mathbf I^{c,1})^{-1}
 \end{equation}
gives

\begin{equation}
\label{eq:momentum_solution2_ore}
\hat{\boldsymbol{\Omega}}_{\Delta}^{c,1}
=
(\epsilon-1)
\mathbf I^{p,i}
(\mathbf I^{p,i}+\mathbf I^{c,1})^{-1}
\mathbf g(
\hat{\boldsymbol{\Omega}}^{p,i}_t,
\hat{\boldsymbol{\Omega}}^{c,1}_t
).
\end{equation}

Finally, the solutions in \eqref{eq:momentum_solution1_ore} and \eqref{eq:momentum_solution2_ore}, together with \eqref{eq:rodriguez}, \eqref{eq:rx1}, and \eqref{eq:rx2}, are used to update all associated vertices.

\subsubsection{Algorithm to Implement Corrections Due To Constraint Enforcement}

As described earlier, we run each of these optimization problems sequentially to generate a correction to the output of the integration method (Line \ref{alg1:residual} in Algorithm \ref{alg:long time horizon}).
This is summarized in Algorithm \ref{alg:constraints enforcement}.
Note that one begins by checking whether any constraint is violated by more than some user-specified parameter $\kappa$.
If any of the constraints are violated by more than $\kappa$, then one applies the correction described in the previous subsections to modify the prediction.
This repeats until all constraints are satisfied.
In our implementation, we set $\epsilon = 0.1$ (Theorem \ref{thm:junctionconstraint}) and $\kappa = 0.02$ (Algorithm \ref{alg:constraints enforcement}).
Algorithm~\ref{alg:constraints enforcement} follows a Gauss--Seidel-style sequential projection scheme. As shown empirically in Appendix~\ref{appendix:constraint_satisfaction}, the corresponding squared constraint residuals remain below \(\kappa=0.02\) throughout a \(5\)~s, \(500\)-step open-loop rollout.
\begin{algorithm}[t]
\caption{Momentum Preserving Constraint Enforcement}
\label{alg:constraints enforcement}
\begin{algorithmic}[1]
\Require
\(\pred{X}{t+1}, \pred{X}{t},
\hat{\boldsymbol{\Omega}}_{t}, \kappa > 0\)

\While{
    \(\exists\,(i,j)\) such that
    \(C^i_j(\hat{\mathbf{X}}_{t+1},\mathbf{0})^2 > \kappa\)
}
    \For{\(i=1\) \textbf{to} \(n-1\)}
        \State
        \(
        \hat{\mathbf{X}}_{t+1}
        =
        \hat{\mathbf{X}}_{t+1}
        +
        D^i_I(\hat{\mathbf{X}}_{t+1})
        \)
        \Comment{Thm.~\ref{theorem:inextensibility}}

        \If{\(i\)th segment is a junction}
            \State
            \(
            \hat{\mathbf{X}}_{t+1}
            =
            \hat{\mathbf{X}}_{t+1}
            +
            D^i_A(\hat{\mathbf{X}}_{t+1})
            \)
            \Comment{Thm.~\ref{theorem:constraint_attachement}}

            \State
            update
            \(
            \hat{\boldsymbol{\Omega}}_{t+1}
            (
                \hat{\mathbf{X}}_{t},
                \hat{\mathbf{X}}_{t+1},
                \hat{\boldsymbol{\Omega}}_{t}
            )
            \)
            \Comment{Thm.~\ref{thm:junctionconstraint}}

            \State
            \(
            \hat{\mathbf{X}}_{t+1}
            =
            \hat{\mathbf{X}}_{t+1}
            +
            D^i_O
            (
                \hat{\mathbf{X}}_{t+1},
                \hat{\boldsymbol{\Omega}}_{t+1})\)
            \Comment{Thm.~\ref{thm:junctionconstraint}}
        \EndIf
    \EndFor
\EndWhile

\State
\Return:
\(
\hat{\mathbf{X}}_{t+1},
\hat{\boldsymbol{\Omega}}_{t+1}
\)

\end{algorithmic}
\end{algorithm}
Because the velocity is recomputed from the corrected positions in Line~7, the mass-weighted zero-sum position corrections preserve total linear momentum.
\subsection{Algorithm Efficiency Improvement}
\label{section:gradientopt}
Modeling speed is critical for efficient planning.
However, as demonstrated in the experiments, directly implementing Algorithm \ref{alg:long time horizon} results in slow simulation and planning.
This section presents two methods to significantly enhance computational performance.

\subsubsection{Analytical Gradient Derivation}

The goal of this subsection is to present the analytical gradient of $P(\MaterialFrame(\mathbf{X}_t, \bm{\theta}_t), \materialp)$ with respect to $\bm{\theta}_t$.
To improve the paper’s readability, we summarize the result in the following theorem.
\begin{thm}
\label{thm:potential_energy_gradient}
Let $\MaterialFrame(\mathbf{X}_t, \bm{\theta}_t)$ be the Material Frame at time $t$, and let
$\materialp$ contain the bending stiffness $\mathbf{B} \in \mathbb{R}^{3\times3} $ and twisting stiffness $\beta \in \mathbb{R}$.
Suppose the total potential energy is the sum of bending and twisting potential energies:
\begin{align}
  P(\MaterialFrame(\mathbf{X}_t, \bm{\theta}_t), \materialp)
  = & P_\text{bend}\bigl(\MaterialFrame(\mathbf{X}_t, \bm{\theta}_t), \mathbf{B}\bigr) +
  \label{eq:potentialP}    \notag \\
  &\hspace{5mm} P_\text{twist}\bigl(\MaterialFrame(\mathbf{X}_t, \bm{\theta}_t), \beta\bigr).
\end{align}
Then the partial derivative of the bending and twisting potential energies with respect to $\theta^i$ are given by
\begin{align}
  \hspace{-3mm}\frac{\partial P_{\mathrm{bend}}(\MaterialFrame(\mathbf{X}_t, \bm{\theta}_t), \materialp)}{\partial \theta^i}
  &= \sum_{k=i}^{\,i+1}
  \bigl(\mathbf{B}^k\, (\bm{\omega}^{(k,i)} - \bar{\bm{\omega}}^{(k,i)})\bigr)^{T}
  \label{eq:potentialbendgradient_paper} \cdot    \notag \\
  &\hspace{20mm}
  \begin{bmatrix}0 & 1\\[6pt]-1 & 0\end{bmatrix}
  \bm{\omega}^{(k,i)},
\end{align}
\begin{align}
  \frac{\partial P_{\mathrm{twist}}(\MaterialFrame(\mathbf{X}_t, \bm{\theta}_t), \materialp)}{\partial \theta^i}
  &= \beta^i \,\bigl(\theta^i - \theta^{i-1}\bigr) \notag \\
  &\hspace{8mm}\quad - \beta^{i+1}\,\bigl(\theta^{i+1} - \theta^i\bigr).
  \label{eq:potentialtwistgradient_paper}
\end{align}
respectively, where the material curvature $\bm{\omega}^{(i,j)}$ is defined as
\begin{equation}
  \hspace{-2mm}\bm{\omega}^{(i,j)} \;=\;
  \bigl(\bcurvature \cdot \mathbf{m}_1^j,\;\bcurvature \cdot \mathbf{m}_2^j\bigr)^T
  \hspace{1.5mm}\text{for}\hspace{1.5mm} j \in \{i-1,\,i\},
  \label{eq:materialcurvature_paper}
\end{equation}
\(\overline{\bm{\omega}}^{(i,j)}\) denotes the undeformed material curvature,
and the curvature vector $\bcurvature$ in \eqref{eq:materialcurvature_paper} is
\begin{equation}
  \kappa \,\mathbf{b}^i
  \;=\;
  \frac{2\,\bigl(\mathbf{e}^{i-1} \times \mathbf{e}^i\bigr)}
  {\|\mathbf{e}^{i-1}\|_2\,\|\mathbf{e}^{i}\|_2 \;+\;\mathbf{e}^{i-1} \cdot \mathbf{e}^{i}}.
  \label{eq:curvature}
\end{equation}
\end{thm}

\textit{Proof of Theorem \ref{thm:potential_energy_gradient}.}
\label{appendix: Theorem 1 Proof}
To prove the theorem, we begin by introducing the curvature binormal and material curvature.

\textbf{Material Curvatures:}
The curvature binormal $\bcurvature$ is traditionally used to represent the turning angle and axis between two consecutive edges:
\begin{equation}
    \kappa \mathbf{b}^i = \frac{2 \, \mathbf{e}^{i-1} \times \mathbf{e}^{i}}{\|\mathbf{e}^{i-1}\|_2 \|\mathbf{e}^{i}\|_2 + \mathbf{e}^{i-1} \cdot \mathbf{e}^{i}},
   \label{eq:curvature_appendix}
\end{equation}
where \(\mathbf{e}^{i-1}\) and \(\mathbf{e}^i\) are consecutive edge vectors, \(\times\) denotes the cross product, and \(\cdot\) denotes the dot product.
By incorporating the material frame \eqref{eq:m1} and \eqref{eq:m2}, the curvature binormal \eqref{eq:curvature_appendix} is projected onto the material frame to quantify the extent to which the curvature aligns with the frame's orientation.
This projection, which we call the deformed material curvature, quantifies the curvature of the current configuration relative to the material frame and is defined as:
\begin{equation}
    \hspace{-2mm}\bm{\omega}^{(i,j)} =
    \left(
        \bcurvature \cdot \mathbf{m}_1^j,
        \bcurvature \cdot \mathbf{m}_2^j
    \right)^T
    \hspace{2mm} \text{for} \hspace{2mm} j \in \{i-1,i\}.
    \label{eq:materialcurvature}
\end{equation}
In \eqref{eq:materialcurvature}, \(\bcurvature\) and
\(\mathbf{m}_1^j,\mathbf{m}_2^j\) are evaluated using the current
configuration, resulting in the deformed material curvature
\(\bm{\omega}^{(i,j)}\). The undeformed material curvature
\(\bar{\bm{\omega}}^{(i,j)}\) is computed using the same definition,
but with the curvature binormal and material frame evaluated in the undeformed configuration.

\textbf{Potential Energy:}
We expand the bending energy and twisting energy in \eqref{eq:potentialP} as
\begin{align}
  P_\text{bend}(\MaterialFrame(\mathbf{X}_t, \bm{\theta}_t), \materialp)
  &= \sum_{i=1}^{n-1} \sum_{j=i-1}^{i} \frac{1}{2}\left( \bm{\omega}^{(i,j)} - \overline{\bm{\omega}}^{(i,j)} \right)^T \notag \\
  &\hspace{15mm} \mathbf{B}^j \left( \bm{\omega}^{(i,j)} - \overline{\bm{\omega}}^{(i,j)} \right),
  \label{eq:potentialPbend}
\end{align}
and
\begin{align}
  P_\text{twist}(\MaterialFrame(\mathbf{X}_t, \bm{\theta}_t), \materialp)
  &= \sum_{i=1}^{n-1} \frac{1}{2} \beta^i \left(\theta^i - \theta^{i-1} \right)^2.
  \label{eq:potentialPtwist}
\end{align}
where \(\overline{\bm{\omega}}^{(i,j)}\) denotes the undeformed material curvature, which is calculated when the DLO is in a static state without any external or internal forces applied, and $\mathbf{B}$ and $\beta$, are components of $\materialp$, representing the bending stiffness and twisting stiffness, respectively.

\textbf{Gradient of Potential Energy}: With \eqref{eq:potentialPbend} and \eqref{eq:potentialPtwist}, the gradient of \eqref{eq:potentialP} can be derived as follows:
\begin{align}
  \frac{\partial P(\MaterialFrame(\mathbf{X}_t, \bm{\theta}_t), \materialp)}{\partial \theta^i}
  &= \frac{\partial P_{bend}(\MaterialFrame(\mathbf{X}_t, \bm{\theta}_t), \materialp)}{\partial \theta^i} + \notag \\
  & \quad \hspace{10mm}\frac{\partial P_{twist}(\MaterialFrame(\mathbf{X}_t, \bm{\theta}_t), \materialp)}{\partial \theta^i}.
  \label{eq:potentialderivative}
\end{align}
We begin by deriving the first term.
Note that $\theta^i$ is only relevant in $\bm{\omega}^{(i,i)}$ and $\bm{\omega}^{(i+1,i)}$.
Using the chain rule, we obtain:
\begin{align}
  \frac{\partial P_{bend}(\MaterialFrame(\mathbf{X}_t, \bm{\theta}_t), \materialp)}{\partial \theta^i}
  &= \sum_{k=i}^{i+1} \frac{\partial P_{bend}(\MaterialFrame(\mathbf{X}_t, \bm{\theta}_t), \materialp)}{\partial \bm{\omega}^{(k,i)}} \cdot \notag \\
  &\hspace{30mm} \frac{\partial \bm{\omega}^{(k,i)}}{\partial \theta^i}.
  \label{eq:bending_theta_derivative}
\end{align}
To compute $\frac{\partial \bm{\omega}^{(k,i)}}{\partial \theta^i}$, we use the identities $\frac{\partial \mathbf{m}_1^i}{\partial \theta^i} = \mathbf{m}_2^i$ and $\frac{\partial \mathbf{m}_2^i}{\partial \theta^i} = -\mathbf{m}_1^i$.
These lead to:
\begin{equation}
    \frac{\partial \bm{\omega}^{(k, i)}}{\partial \theta^i} = \begin{bmatrix}
0 & 1 \\
-1 & 0
\end{bmatrix} \bm{\omega}^{(k, i)}
\end{equation}
Substituting this result back into \eqref{eq:bending_theta_derivative} yields \eqref{eq:potentialbendgradient_paper}.
The second term of \eqref{eq:potentialderivative} follows from differentiating \eqref{eq:potentialPtwist} with respect to $\theta$ and results in \eqref{eq:potentialtwistgradient_paper}.

\subsubsection{Parallel Programming.}
Unlike quasi-static approaches \citep{PBD_DER, baseline1} that constantly require handling interactions between parent and child branches at each time step, DEFT constructs a computational graph that computes dynamics at junctions through constraints as post-processing.
This strategy enables efficient parallel execution of Algorithm \ref{alg:long time horizon} (Lines 1-5 and 7) via PyTorch's batch-wise operations, seamlessly integrating the physics-based framework with residual learning methods.
To address dimensional mismatches between parent ($n^p$) and child ($n^c$) branches, we pad child branch matrices and vertices with zeros, ensuring uniform batch dimensions.
Additionally, gradients from padded elements are masked during parameter updates to avoid influencing learned parameters.
\section{DEFT for Trajectory Optimization}
\label{section:planning}

DEFT’s differentiable forward model enables gradient-based trajectory optimization for BDLO manipulation.
In this section, we present a general optimization framework that leverages DEFT’s predictive capability and differentiability, followed by two example applications: shape matching and thread insertion.
We then detail the implementation, clarify the level of autonomy, and discuss potential directions toward achieving full autonomy.

\subsection{General Framework}

Given the initial state of the BDLO,
$\mathbf{X}_{t=1}$, $\mathbf{V}_{t=1}$, and $\boldsymbol{\Omega}_{t=1}$,
and a sequence of control inputs $\{\mathbf{u}_t\}_{t=1}^{T-1}$,
we formulate the following trajectory optimization problem over a finite time horizon:
\begin{align}
\min_{\{\mathbf{u}_t\}_{t=1}^{T-1}}
\quad &
\mathcal{J}(\hat{\mathbf{X}}_{T}) \notag \\
\text{s.t.} \quad &
(\hat{\mathbf{X}}_{t+1}, \hat{\mathbf{V}}_{t+1}, \hat{\boldsymbol{\Omega}}_{t+1})
=
\text{DEFT}(\hat{\mathbf{X}}_{t}, \hat{\mathbf{V}}_{t}, \hat{\boldsymbol{\Omega}}_{t}, \mathbf{u}_t), \notag \\
& \forall t = 1,\dots,T-1.
\label{eq:single_shooting_full}
\end{align}
Here, $\mathcal{J}$ denotes a task-specific objective evaluated at the terminal state.
We adopt a direct single-shooting formulation~\citep{single-shooting}, where the state trajectory is obtained via forward rollout of the DEFT dynamics, resulting in dynamic feasibility by construction.
The optimization is performed over the control sequence, with gradients computed by backpropagating through the differentiable DEFT rollout.
This involves differentiating the dynamics with respect to both control inputs and state variables at each time step, which is handled via automatic differentiation in PyTorch~\citep{pytorch}.

\subsection{Applications}

\subsubsection{Shape Matching}

We instantiate \eqref{eq:single_shooting_full} for a shape-matching task,
where a subset of node indices
$\mathcal{S} \subseteq \{1, \dots, n\}$ is specified along with their desired target positions
$\{\mathbf{x}^{i,\mathrm{target}}_{T}\}_{i \in \mathcal{S}}$.
The objective is defined as
\begin{equation}
\mathcal{J}_{\mathrm{shape}} =
\sum_{i \in \mathcal{S}}
\left\|
\hat{\mathbf{x}}^{i}_{T} - \mathbf{x}^{i,\mathrm{target}}_{T}
\right\|_2^2.
\end{equation}
Since the objective depends only on the terminal state, we parameterize the control sequence using a low-dimensional representation (e.g., a single control variable $\mathbf{u}_{T-1}$),
with intermediate controls generated via linear interpolation between $\mathbf{u}_1$ and $\mathbf{u}_{T-1}$.
The trajectory is then obtained via forward rollout of the DEFT dynamics.

\subsubsection{Thread Insertion}

The goal of this task is to guide the tip of the BDLO through a target hole.
Let $i_{\mathrm{tip}}$ denote the index of the tip vertex.
Given the pose of the target hole, we denote its center as $\mathbf{x}^{\mathrm{hole}} \in \mathbb{R}^3$ and the normal direction as $\mathbf{n} \in \mathbb{R}^3$.
A key challenge is that directly performing insertion may cause the motion of the BDLO to be obstructed by the boundary of the hole, resulting in failure.
To address this, we define staged waypoint targets along the normal direction:
\begin{equation}
\label{eq:insert_wp}
\mathbf{x}^{\mathrm{wp}} =
\begin{cases}
\mathbf{x}^{\mathrm{hole}} + \delta_{\mathrm{pre}} \mathbf{n}, & \text{pre-insertion}, \\
\mathbf{x}^{\mathrm{hole}} + \delta_{\mathrm{ins}} \mathbf{n}, & \text{insertion},
\end{cases}
\end{equation}
where $\delta_{\mathrm{pre}} > 0$ places the tip in front of the hole to guide alignment with the central opening, and $\delta_{\mathrm{ins}} < 0$ drives the tip through the hole.
The objective is defined as
\begin{equation}
\mathcal{J}_{\mathrm{insert}} =
\left\|
\hat{\mathbf{x}}^{i_{\mathrm{tip}}}_{T} - \mathbf{x}^{\mathrm{wp}}
\right\|_2^2.
\end{equation}

Similar to the shape-matching task, we parameterize the control sequence using a low-dimensional representation, where intermediate controls are generated via linear interpolation.
In practice, we solve two sequential optimization problems corresponding to the pre-insertion and insertion stages, resulting in a staged waypoint strategy within the same trajectory optimization framework.

\subsection{Setup and Implementation}
Because this work focuses on modeling and trajectory optimization, the BDLO state and the pose of the target hole are obtained using motion capture.
Specifically, for shape matching, motion capture (MoCap) markers are attached to the BDLO to obtain the initial state $\mathbf{X}_{t=1}$ and target shape.
For insertion tasks, only the initial state is obtained using markers.
Three markers placed on the target hole form an approximately equilateral triangle, enabling estimation of the hole center $\mathbf{x}^{\mathrm{hole}}$ and its normal direction $\mathbf{n}$.
Waypoints are then generated using \eqref{eq:insert_wp}. The perception component can be replaced by existing methods, such as markerless tracking for BDLO state estimation~\citep{trackdeform3d} and pose estimation methods for the target hole~\citep{gsposeest}.

Given these estimated states, the control sequence $\{\mathbf{u}_t\}_{t=1}^{T-1}$ is obtained by solving \eqref{eq:single_shooting_full} using the proposed DEFT-based trajectory optimization framework, with all trajectory generation and control computed autonomously.
No pre-recorded trajectories or manually designed waypoints are used.
All waypoints are generated automatically from the task specification.
We solve \eqref{eq:single_shooting_full} using IPOPT~\citep{ipopt} via its Python interface~\citep{cyipopt}.
\section{Experiments and Results}

\begin{figure}[t]
    \centering
    \includegraphics[width=0.5\textwidth]{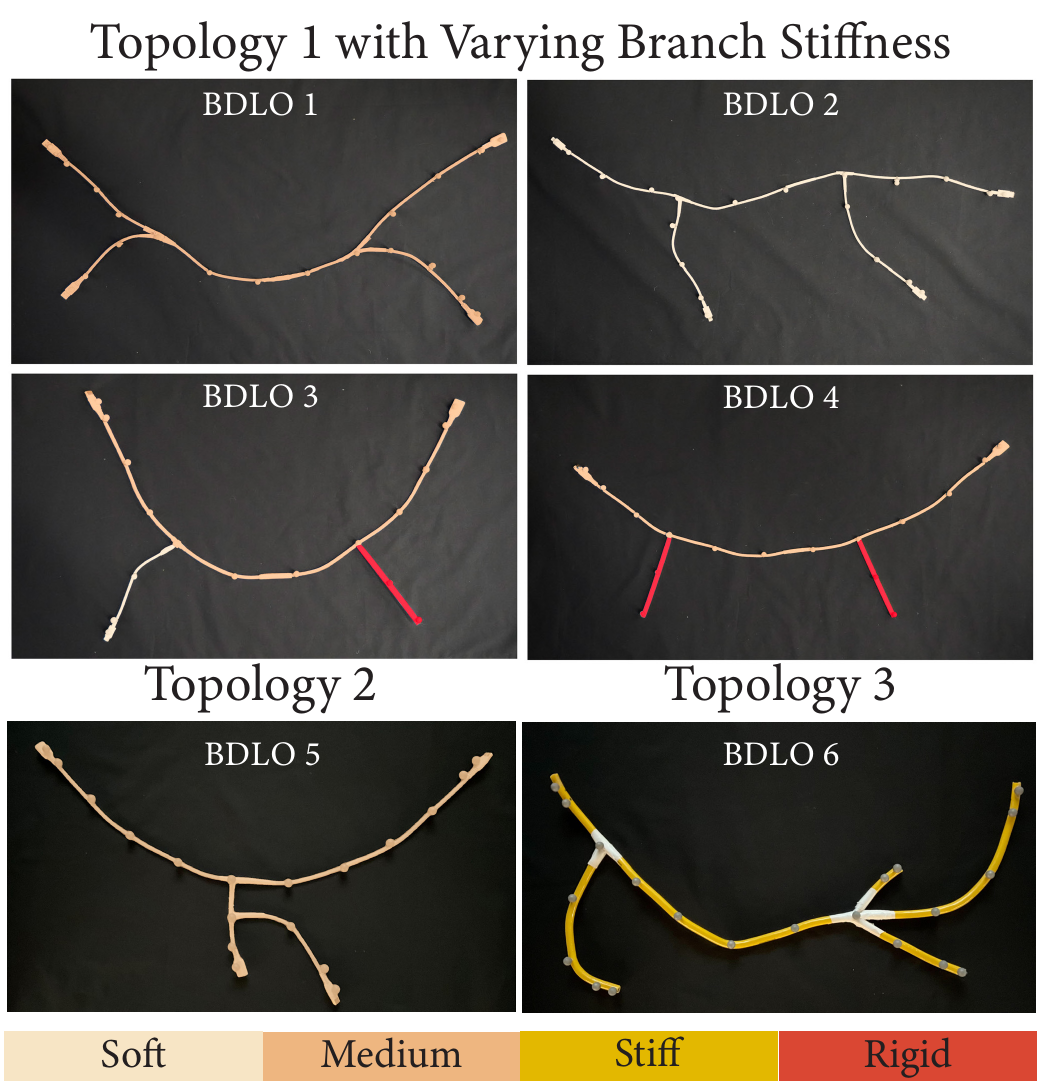}
    \caption{An illustration of the BDLOs used to evaluate and compare DEFT against baseline methods.
To enhance dataset diversity, we vary stiffness (measured on a relative scale), length, mass, number of vertices, and junction topology.
\textit{In real-world wire harnesses, rigid components such as clamps and fixation elements are commonly attached along the wires~\citep{wireharnesinstruction}. To reflect this, we construct BDLOs that integrate deformable segments with rigid components modeled as high-stiffness regions. This setup highlights DEFT’s ability to simultaneously capture deformable behavior and interactions with rigid structures.}}
    \label{fig:bdlos}
\end{figure}

\begin{table*}[t]
\centering
\caption{Topology and branch-wise properties of each BDLO used in the experiments.
Topology configurations are illustrated in Figure~\ref{fig:bdlos}.
Branch stiffness is defined on a relative scale with four levels: soft (S), medium (M), stiff (H), and rigid (R).
Dataset durations for each BDLO are reported in the final two rows.}
\label{mat_prop_img}
\begin{tabular}{c||c c c c c c}
\toprule

& BDLO 1 & BDLO 2 & BDLO 3 & BDLO 4 & BDLO 5 & BDLO 6 \\

\midrule

Topology
& 1
& 1
& 1
& 1
& 2
& 3 \\

\midrule

\makecell{Branch Stiffness}
& $[M,\; M,\; M]$
& $[S,\; S,\; S]$
& $[M,\; S,\; R]$
& $[M,\; R,\; R]$
& $[M,\; M,\; M]$
& $[H,\; H,\; H]$ \\

\makecell{Marker Count}
& $[13,\; 4,\; 3]$
& $[12,\; 4,\; 4]$
& $[12,\; 2,\; 3]$
& $[12,\; 2,\; 2]$
& $[12,\; 3,\; 3]$
& $[12,\; 4,\; 3, \;2]$ \\

\midrule

Dataset 1 (s)
& 505 & 735 & 645 & 495 & 505 & 435 \\

Dataset 2 (s)
& 440 & N/A & 445 & N/A & N/A & N/A \\

\bottomrule
\end{tabular}
\label{tab:bdlo_prop}
\end{table*}
\label{sec:experiments}
This section evaluates DEFT through real-world experiments:
1) A quantitative accuracy comparison with state-of-the-art methods, including scenarios where grasping occurs at various points along the BDLO, such as parent-branch ends and midpoints;
2) An assessment of DEFT’s inference speed;
3) An ablation study exploring parameter auto-tuning, residual learning, and a naive multi-branch approach that connects multiple single-DLO models;
4) Shape-matching manipulation demonstrations, highlighting DEFT’s real-world manipulation efficacy;
5) Thread insertion manipulation demonstrations, showcasing DEFT’s versatility.
The source code and data are available at https://github.com/roahmlab/DEFT.
To the best of our knowledge, this will be the first publicly available dataset and code for modeling BDLOs.

\subsection{Hardware Setup}
\begin{figure}[t]
    \centering
    \includegraphics[width=0.5\textwidth]{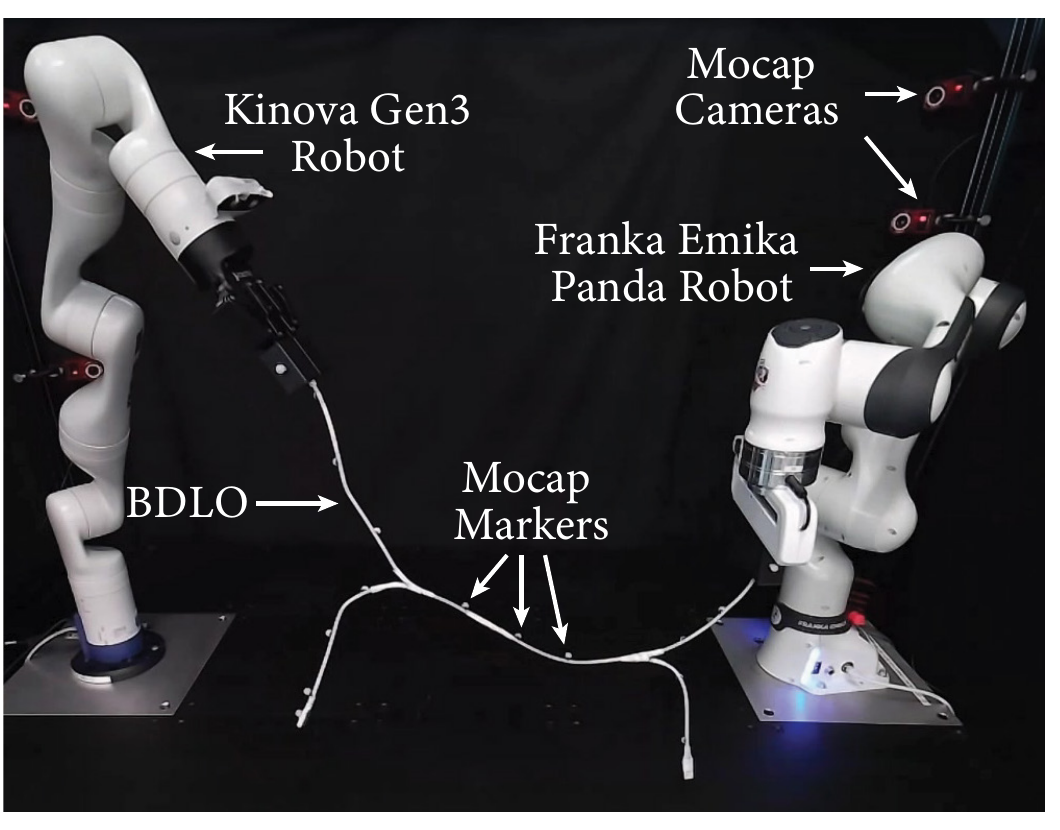}
    \caption{Experimental setup for data collection and manipulation using a dual-arm robotic system.}
    \label{fig:experimentsetup}
\end{figure}
Six distinct BDLOs were constructed to validate DEFT’s modeling accuracy, as illustrated in Figure~\ref{fig:bdlos}.
Four of these share the same topology but differ in the stiffness of each branch. Notably, in BDLO3 and BDLO4, we introduce stiff child branches attached to the parent branch. These branches represent rigid components (e.g., fixations) commonly used in BDLO construction~\citep{wireharnesinstruction}. Such attachments help ensure secure mounting, preserve structural integrity, and manage loads in real-world applications.
The remaining two BDLOs exhibit distinct topologies. BDLO5 includes a second furcation along a child branch, while BDLO6 contains a junction with four segments.
Both configurations are representative of structures commonly encountered in real-world applications.
Spherical markers were attached to each BDLO to facilitate dataset generation for both training and evaluation, with the number of vertices set equal to the number of MoCap markers. Since the markers are present during both training and evaluation, their influence on the BDLO dynamics is inherently captured in all subsequent analyses. It is important to note that these markers are \emph{only} used for dataset construction and performance evaluation of \DEFTn through ground-truth measurements.
For real-world deployment, markers are not required. Markerless tracking methods, such as~\citep{trackdeform3d}, can be used to construct BDLO datasets without motion capture. While such approaches provide a viable offline data acquisition pipeline, they are not directly suitable for real-time robotic manipulation, as markerless tracking alone does not provide a BDLO model.
For dual-arm manipulation, we employ a Franka Emika Research 3 robot and a Kinova Gen3 robot.
Ground-truth vertex positions are recorded using an OptiTrack motion capture system.
An overview of the experimental setup is shown in Figure~\ref{fig:experimentsetup}.

\begin{table*}[t]

\centering
\caption{\textbf{Left:} Modeling results over a \(5\)s horizon.
For learned methods, each RMSE entry reports the mean on the first line
and the standard deviation on the second line, computed across evaluation
trajectories and three independently trained seeds.
\textbf{Right:} Computational time per inference step
(\(10^{-2}\,\mathrm{s}\)). For learned methods, the mean and standard
deviation across the three seeds are reported using the same format.}
\label{tab:model_accuracy}

\small
\begingroup
\setlength{\tabcolsep}{2.5pt}
\renewcommand{\arraystretch}{1.10}

\begin{tabular}{@{}l||cccccc|cc||c@{}}
    \toprule
    & \multicolumn{8}{c||}{
        Modeling Accuracy (RMSE, \(10^{-2}\,\mathrm{m}\))
    }
    & \multicolumn{1}{c}{Time} \\

    & \multicolumn{6}{c|}{Dataset 1}
    & \multicolumn{2}{c||}{Dataset 2}
    & \multicolumn{1}{c}{(\(10^{-2}\,\mathrm{s}\))} \\

    Method & 1 & 2 & 3 & 4 & 5 & 6 & 1 & 3 & 1 \\
    \hline

    GCN~\citep{GCN}
    & \shortstack{6.64\\[-2pt]{\footnotesize\(\pm\,0.20\)}}
    & \shortstack{6.52\\[-2pt]{\footnotesize\(\pm\,0.43\)}}
    & \shortstack{5.41\\[-2pt]{\footnotesize\(\pm\,0.15\)}}
    & \shortstack{4.30\\[-2pt]{\footnotesize\(\pm\,0.11\)}}
    & \shortstack{4.50\\[-2pt]
        {\footnotesize\(\pm\,0.30\)}}
    & \shortstack{6.82\\[-2pt]
        {\footnotesize\(\pm\,0.04\)}}
    & \shortstack{7.29\\[-2pt]{\footnotesize\(\pm\,0.14\)}}
    & \shortstack{4.35\\[-2pt]{\footnotesize\(\pm\,0.18\)}}
    & \cellcolor{red!30}
      \shortstack{0.163\\[-2pt]
        {\footnotesize\(\pm\,0.01\)}} \\

    \specialrule{0.25pt}{1pt}{1pt}

    Tree-LSTM~\citep{treeLSTM}
    & \shortstack{5.32\\[-2pt]{\footnotesize\(\pm\,0.12\)}}
    & \shortstack{4.45\\[-2pt]{\footnotesize\(\pm\,0.20\)}}
    & \shortstack{2.50\\[-2pt]{\footnotesize\(\pm\,0.08\)}}
    & \shortstack{2.82\\[-2pt]{\footnotesize\(\pm\,0.13\)}}
    & \shortstack{5.06\\[-2pt]
        {\footnotesize\(\pm\,0.07\)}}
    & \shortstack{4.96\\[-2pt]
        {\footnotesize\(\pm\,0.10\)}}
    & \shortstack{7.13\\[-2pt]{\footnotesize\(\pm\,0.09\)}}
    & \shortstack{4.54\\[-2pt]{\footnotesize\(\pm\,0.14\)}}
    & \shortstack{2.02\\[-2pt]
        {\footnotesize\(\pm\,0.26\)}} \\

    \specialrule{0.25pt}{1pt}{1pt}

    FEM~\citep{FEM}: \(\Delta_t\): \(10^{-6}\)s, node: 60
    & \cellcolor{yellow!30}3.11
    & \cellcolor{yellow!30}3.70
    & \cellcolor{yellow!30}2.29
    & \cellcolor{yellow!30}2.84
    & \cellcolor{yellow!30}2.97
    & \cellcolor{yellow!30}3.42
    & \cellcolor{yellow!30}2.17
    & \cellcolor{yellow!30}2.10
    & 493 \\

    \specialrule{0.25pt}{1pt}{1pt}

    FEM~\citep{FEM}: \(\Delta_t\): \(10^{-6}\)s, node: 160
    & \cellcolor{orange!30}2.07
    & \cellcolor{orange!30}3.35
    & \cellcolor{orange!30}1.82
    & \cellcolor{orange!30}1.54
    & \cellcolor{orange!30}2.13
    & \cellcolor{orange!30}2.10
    & \cellcolor{orange!30}1.40
    & \cellcolor{orange!30}1.94
    & 1211 \\

    \specialrule{0.25pt}{1pt}{1pt}

    FEM~\citep{FEM}: \(\Delta_t\): \(10^{-2}\)s, node: 60
    & 14.6
    & 18.3
    & 12.1
    & 17.6
    & 26.2
    & 18.9
    & 14.9
    & 12.3
    & \cellcolor{orange!30}0.485 \\

    \specialrule{0.25pt}{1pt}{1pt}

    FEM~\citep{FEM}: \(\Delta_t\): \(10^{-2}\)s, node: 160
    & 12.9
    & 15.1
    & 10.0
    & 11.8
    & 13.5
    & 16.8
    & 12.8
    & 9.71
    & 1.09 \\

    \specialrule{0.25pt}{1pt}{1pt}

    AdaptiGraph~\citep{AdaptiGraph}
    & \shortstack{4.01\\[-2pt]{\footnotesize\(\pm\,0.11\)}}
    & \shortstack{4.45\\[-2pt]{\footnotesize\(\pm\,0.14\)}}
    & \shortstack{3.08\\[-2pt]{\footnotesize\(\pm\,0.38\)}}
    & \shortstack{2.69\\[-2pt]{\footnotesize\(\pm\,0.11\)}}
    & \shortstack{2.82\\[-2pt]
        {\footnotesize\(\pm\,0.06\)}}
    & \shortstack{3.84\\[-2pt]
        {\footnotesize\(\pm\,0.04\)}}
    & \shortstack{2.25\\[-2pt]{\footnotesize\(\pm\,0.08\)}}
    & \shortstack{2.53\\[-2pt]{\footnotesize\(\pm\,0.12\)}}
    & \shortstack{0.168\\[-2pt]
        {\footnotesize\(\pm\,0.0\)}} \\

    \specialrule{0.25pt}{1pt}{1pt}

    DEFORM\(^{+}\)~\citep{DEFORM}
    & \shortstack{4.57\\[-2pt]{\footnotesize\(\pm\,0.06\)}}
    & \shortstack{7.51\\[-2pt]{\footnotesize\(\pm\,0.08\)}}
    & \shortstack{5.04\\[-2pt]{\footnotesize\(\pm\,0.05\)}}
    & \shortstack{4.38\\[-2pt]{\footnotesize\(\pm\,0.15\)}}
    & \shortstack{3.33\\[-2pt]
        {\footnotesize\(\pm\,0.06\)}}
    & \shortstack{5.78\\[-2pt]
        {\footnotesize\(\pm\,0.18\)}}
    & \shortstack{3.91\\[-2pt]{\footnotesize\(\pm\,0.07\)}}
    & \shortstack{4.36\\[-2pt]{\footnotesize\(\pm\,0.13\)}}
    & \shortstack{1.665\\[-2pt]
        {\footnotesize\(\pm\,0.03\)}} \\

    \specialrule{0.25pt}{1pt}{1pt}

    DEFT
    & \cellcolor{red!30}
      \shortstack{1.44\\[-2pt]{\footnotesize\(\pm\,0.07\)}}
    & \cellcolor{red!30}
      \shortstack{2.14\\[-2pt]{\footnotesize\(\pm\,0.04\)}}
    & \cellcolor{red!30}
      \shortstack{1.20\\[-2pt]{\footnotesize\(\pm\,0.04\)}}
    & \cellcolor{red!30}
      \shortstack{0.93\\[-2pt]{\footnotesize\(\pm\,0.09\)}}
    & \cellcolor{red!30}
      \shortstack{0.96\\[-2pt]
        {\footnotesize\(\pm\,0.01\)}}
    & \cellcolor{red!30}
      \shortstack{1.15\\[-2pt]
        {\footnotesize\(\pm\,0.08\)}}
    & \cellcolor{red!30}
      \shortstack{1.08\\[-2pt]{\footnotesize\(\pm\,0.13\)}}
    & \cellcolor{red!30}
      \shortstack{1.07\\[-2pt]{\footnotesize\(\pm\,0.02\)}}
    & \cellcolor{yellow!30}
      \shortstack{0.825\\[-2pt]
        {\footnotesize\(\pm\,0.03\)}} \\

    \bottomrule
\end{tabular}

\endgroup
\end{table*}
\subsection{Dataset Collection and Training}

\subsubsection{Dataset Collection}
We collected dynamic trajectory data for selected BDLOs in the real world using a motion capture system operating at 100 Hz, capturing both slow and fast BDLO motions, as illustrated at the top of Figure \ref{fig:experimentsetup}.
We split our data collection into two categories: Dataset 1. each robot grasps one end of the BDLO.  Data from this category were collected for all BDLOs.
Dataset 2. One robot grasped one end of the BDLO while the other robot grasped the BDLO at its midpoint.
This dataset is collected for BDLOs 1 and 3.
Total collection times are provided in Table \ref{tab:bdlo_prop}.
The dataset is split with \(\sim \)75\% for training and \(\sim \)25\% for evaluation.

\subsubsection{Training Setup}
\label{Training Details}
The model for each of the six BDLOs is trained separately using the trajectory data collected from our real-world experiments.
During training, models are trained using the same 50 steps (i.e., 0.5s) of dynamic motion.
Each method's performance is evaluated when predicting the state of the wire over a 500 step (i.e. 5s) prediction horizon using just the state of the wire at the initial time and the orientation and position of the two end effectors at each time step.
Let $\mathbf{U}_{1:T-1}$ denote the set of inputs applied from time $t=1$ to $t=T-1$, and let $\mathbf{X}_{1:T} = \{\mathbf{X}_1, \mathbf{X}_2, \dots, \mathbf{X}_T\} \in \mathbb{R}^{T \times n \times 3}$ denote the corresponding ground-truth trajectory of the BDLO from $t=1$ to $t=T$. Given known initial conditions $\mathbf{X}_1$, $\mathbf{V}_1$, and inputs $\mathbf{U}_{1:T-1}$, Algorithm~\ref{alg:long time horizon} can be recursively applied to predict the trajectory $\hat{\mathbf{X}}_{2:T}$. Let $\bm{\phi}$ denote the parameters of the GNN. The training objective is formulated as the following optimization problem:
\begin{equation}
    \label{eq:trainingcost}
    \underset{\bm{\alpha}, \bm{\phi}}{\min} \sum_{t=1}^{T-1} \|\mathbf{X}_{t+1} - \hat{\mathbf{X}}_{t+1}\|_2
\end{equation}

\subsection{Baseline Comparisons}
\subsubsection{Baselines}
 Only a limited number of papers have investigated the dynamic behavior of BDLOs, and none provide a differentiable framework for learning real-world dynamics from real-world datasets.
Moreover, no public code repositories are available for these approaches, making direct replication infeasible.
We select four implementations for comparison.
Tree-LSTM and GCN can both capture structured relationships.
Specifically, Tree-LSTM leverages hierarchical tree representations, while GCN excels in graph-based modeling.
Because BDLOs can be described by nodes and edges, these architectures are naturally able to incorporate a BDLO's dynamics.
We selected FEM~\citep{FEM} as a baseline and optimized its parameters via grid search.
Because FEM simulation is computationally expensive, we do not apply iterative gradient-based identification. Instead, we evaluate a grid of candidate material parameters on the same real-world training portion used by DEFT and select the best-performing parameter set.
Simulations varied timestep sizes from \(1\times10^{-2}\)~s (DEFT's setting) to \(1\times10^{-6}\)~s and used \(60\) or \(160\) nodes, compared with approximately \(20\) nodes for DEFT.
The \(60\)- and \(160\)-node counts refer to the internal FEM simulation resolution. For computing RMSE, the FEM outputs are sampled at regular intervals to obtain the same number of evaluation vertices as DEFT. Accordingly, the FEM results provide a speed--accuracy comparison rather than a controlled comparison isolating only model class or simulation resolution.
We use the highest-resolution FEM simulation to generate 5-minute sequences of simulation data for each BDLO to pretrain AdaptiveGraph~\citep{AdaptiGraph}, and then fine-tune the pretrained model on the real-world dataset.
To highlight DEFT's contributions, we created DEFORM$^+$, which extends DEFORM~\citep{DEFORM} to BDLOs by replacing DEFT’s junction constraints (Section \ref{section:constratins}) with FEM-based springs and bending. DEFORM$^+$ excludes DEFT's residual learning architecture (Section \ref{section:residual learning}) and DEFT’s computational optimizations (Section \ref{section:gradientopt}).
This is the most straightforward extension of DEFORM to BDLOs.
All baselines' training procedures are identical to DEFT’s.

\subsubsection{Modeling Results}
\begin{figure*}[t]
    \centering
    \includegraphics[width=0.82
    \textwidth]{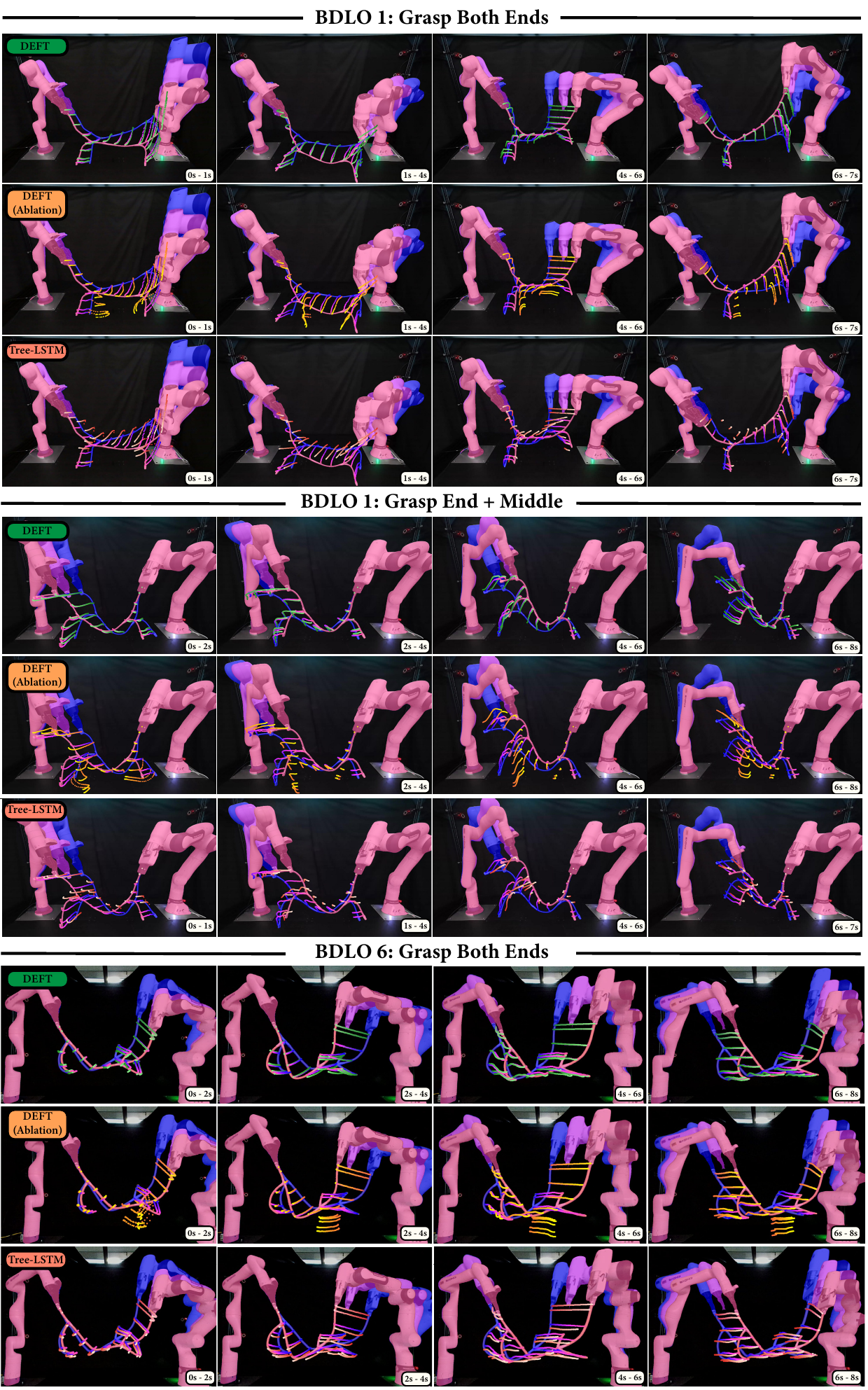}
    \caption{Visualization of the predicted trajectories using DEFT, a DEFT ablation that leaves out the constraint described in Theorem \ref{thm:junctionconstraint}, and Tree-LSTM.
    The ground-truth initial position of the vertices are colored in blue, the ground-truth final position of the vertices are colored in pink, and the gradient between these two colors is used to denote the ground truth location over time.
    The predicted vertices are colored as green circles (DEFT), orange circles (DEFT ablation), and light red circles (Tree-LSTM), respectively.
    A gradient is used for these predictions to depict the evolution of time, starting from dark and going to light.
    Note that the ground truth is only provided at t=0 s.
    The prediction is performed recursively without requiring additional ground-truth data or perception inputs throughout the process.
    Additional visualizations of BDLO~5 can be found in the Appendix~\ref{app_sec:modeling_vis}, Figure~\ref{app_fig:modeling_vis}.}
    \label{fig:modeling_vis}
\end{figure*}

Table~\ref{tab:model_accuracy} (\textbf{left}) shows the average RMSE across two dataset categories over a 5-second prediction horizon, where DEFT outperforms all baselines.
These results demonstrate the ability of DEFT to accurately capture the behavior of BDLOs with diverse topologies and material properties.
Figure~\ref{fig:modeling_vis} visualizes performance comparisons among DEFT, Tree-LSTM, and a DEFT ablation described later; additional visualizations for BDLO~5 are shown in Figure~\ref{app_fig:modeling_vis} in the Appendix.
Experimental videos are available in the supplementary material.
\FloatBarrier

\subsection{Computational Speed}
\label{sec:Computational Speed}
This section evaluates the computational efficiency of DEFT when compared to baseline methods.
It also examines how employing the analytical gradient derived from Theorem \ref{thm:potential_energy_gradient} to compute \eqref{eq:innerloop} and the parallel programming approach detailed in Section \ref{section:gradientopt}, impacts overall computational time.
All experiments in this section were conducted in Python on an Ubuntu 20.04 equipped with an AMD Ryzen PRO 5995WX CPU, 256GB of RAM, and 128 cores.

\subsubsection{Computational Speed Results}
Table~\ref{tab:model_accuracy} (\textbf{right}) summarizes average computational time for 0.01s inference.
Table \ref{tab:model_accuracy} indicates that GCN has the fastest speed.
DEFT is the third fastest and is able to generate single step predictions above 100Hz frequency.
In the planning task described later, the shape-matching planning times are approximately \(8.48\)~s for DEFT, \(73.0\)~s for DEFORM\({}^{+}\), and \(1.62\times10^{4}\)~s (\(4.50\)~h) for the finest FEM baseline.
To evaluate the effect of the available CPU resources, we additionally restrict execution to \(1\), \(8\), and \(16\) logical CPU threads. The corresponding DEFT planning times are \(21.0\), approximately \(8.6\), and \(8.37\)~s, respectively, compared with \(8.48\)~s under unrestricted execution. All configurations follow the same IPOPT convergence path and produce the same successful solution. These results demonstrate an approximately \(2.5\times\) speedup from one to sixteen available CPU threads, with performance saturating at approximately eight to sixteen threads because of the limited branch count and the sequential components of the planning loop.
\begin{figure}[H]
    \centering
    \vspace{-3mm}
    \includegraphics[width=0.45\textwidth]{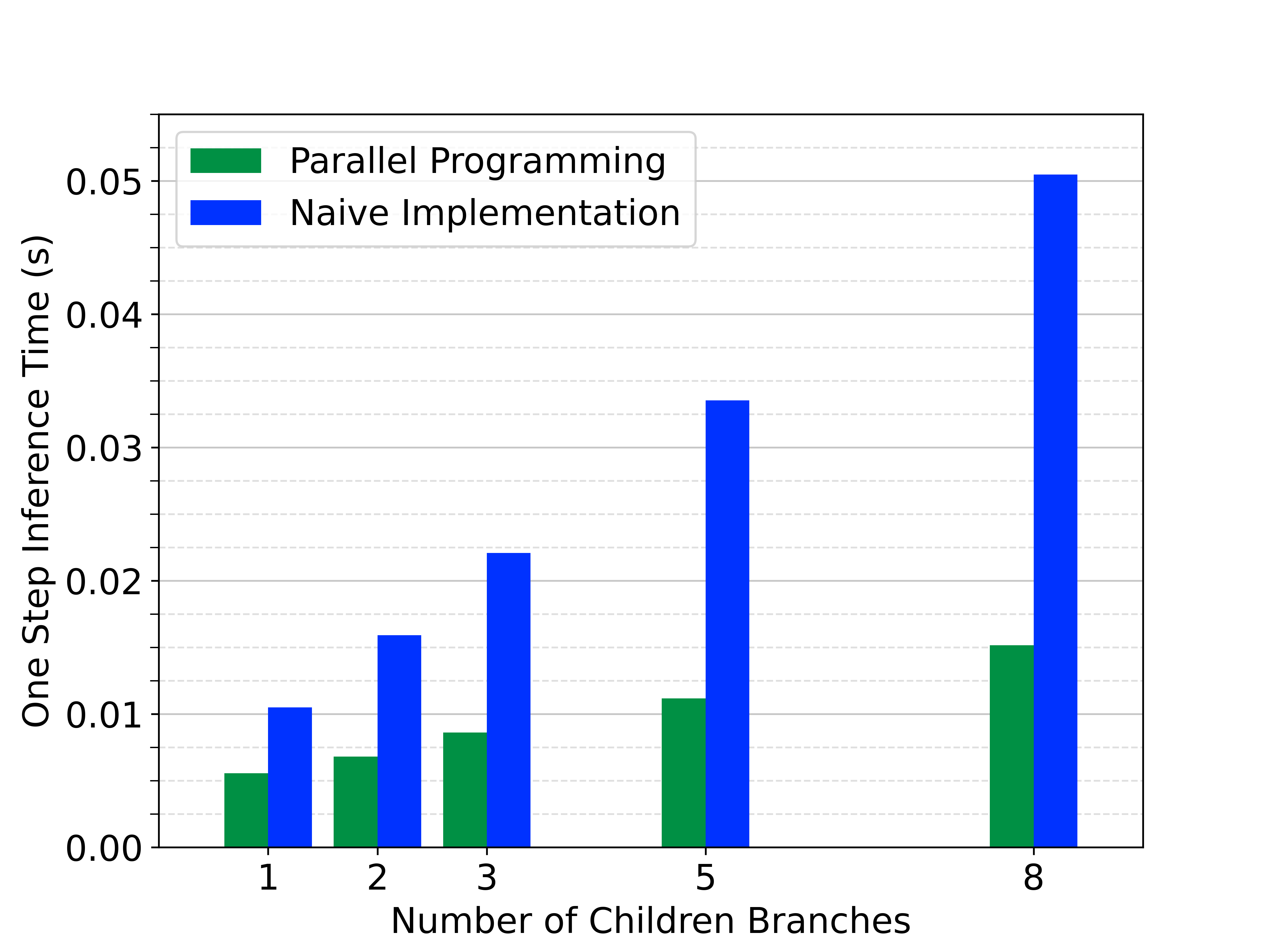}
    \caption{Computational speed comparison: parallel programming versus a naive (sequential) implementation.}
    \label{fig:computational_speed3}
\end{figure}
\begin{table}[t]

    \centering
    \caption{Computational time for \eqref{eq:innerloop} with Dataset~1.
    Each entry reports the mean on the first line and the standard
    deviation on the second line across three timing runs.}
    \begin{tabular}{l||cccc}
        \toprule
        & \multicolumn{4}{c}{Time (\(10^{-2}\,\mathrm{s}\))} \\
        Method & 1 & 2 & 3 & 4 \\
        \hline

        Analytical Gradient
        & \cellcolor{red!30}
          \shortstack{0.34\\[-2pt]
          {\footnotesize\(\pm\,0.02\)}}
        & \cellcolor{red!30}
          \shortstack{0.27\\[-2pt]
          {\footnotesize\(\pm\,0.01\)}}
        & \cellcolor{red!30}
          \shortstack{0.41\\[-2pt]
          {\footnotesize\(\pm\,0.01\)}}
        & \cellcolor{red!30}
          \shortstack{0.46\\[-2pt]
          {\footnotesize\(\pm\,0.01\)}} \\

        \specialrule{0.25pt}{1pt}{1pt}

        Numerical Gradient
        & \cellcolor{orange!30}
          \shortstack{0.74\\[-2pt]
          {\footnotesize\(\pm\,0.02\)}}
        & \cellcolor{orange!30}
          \shortstack{0.65\\[-2pt]
          {\footnotesize\(\pm\,0.01\)}}
        & \cellcolor{orange!30}
          \shortstack{0.84\\[-2pt]
          {\footnotesize\(\pm\,0.04\)}}
        & \cellcolor{orange!30}
          \shortstack{0.91\\[-2pt]
          {\footnotesize\(\pm\,0.02\)}} \\

        \bottomrule
    \end{tabular}
    \label{tab:computational speed2}
\end{table}
We compare the computational speed of using analytical gradients when compared to using numerical gradients.
As summarized in Table \ref{tab:computational speed2}, computing \eqref{eq:innerloop} with the analytical gradient is approximately twice as fast as using numerical gradients.

\subsection{Ablation Study}
\begin{table*}[t]

    \centering
    \caption{Ablation study with Dataset~1 and 2.
    For configurations containing a learned residual model, each cell
    reports the mean RMSE on the first line and the standard deviation
    on the second line, computed across evaluation trajectories and
    three independently trained seeds.}
    \small
    \begin{tabular}{l||cccc|cc}
        \toprule
        & \multicolumn{6}{c}{
            Modeling Accuracy (RMSE, \(10^{-2}\,\mathrm{m}\))\(\downarrow\)
        } \\
        & \multicolumn{4}{c|}{Dataset 1}
        & \multicolumn{2}{c}{Dataset 2} \\
        Method & 1 & 2 & 3 & 4 & 1 & 3 \\
        \midrule

        W/O Constraints.\ref{theorem:inextensibility},
        \ref{theorem:constraint_attachement} and
        \ref{thm:junctionconstraint}
        & \(7.66\times10^8\)
        & \(6.10\times10^8\)
        & \(6.56\times10^6\)
        & \(1.77\times10^8\)
        & \(3.81\times10^7\)
        & \(1.23\times10^8\) \\

        \specialrule{0.25pt}{1pt}{1pt}

        W/O Constraints.\ref{theorem:constraint_attachement} and
        \ref{thm:junctionconstraint}
        & \(1.36\times10^2\)
        & \(1.81\times10^2\)
        & \(1.51\times10^2\)
        & \(7.64\times10^1\)
        & \(8.72\times10^1\)
        & \(3.38\times10^2\) \\

        \specialrule{0.25pt}{1pt}{1pt}

        W/O Constraint.\ref{thm:junctionconstraint}
        & \shortstack{3.95\\[-2pt]
          {\footnotesize\(\pm\,0.23\)}}
        & \shortstack{3.49\\[-2pt]
          {\footnotesize\(\pm\,0.14\)}}
        & \shortstack{2.03\\[-2pt]
          {\footnotesize\(\pm\,0.19\)}}
        & \shortstack{2.15\\[-2pt]
          {\footnotesize\(\pm\,0.08\)}}
        & \shortstack{4.01\\[-2pt]
          {\footnotesize\(\pm\,0.16\)}}
        & \shortstack{4.88\\[-2pt]
          {\footnotesize\(\pm\,0.25\)}} \\

        \specialrule{0.25pt}{1pt}{1pt}

        W/O Learning + System ID
        & 2.46
        & 3.19
        & 1.83
        & 1.81
        & 1.85
        & 2.43 \\

        \specialrule{0.25pt}{1pt}{1pt}

        W/O System ID
        & \cellcolor{orange!30}
          \shortstack{1.75\\[-2pt]
          {\footnotesize\(\pm\,0.11\)}}
        & \cellcolor{orange!30}
          \shortstack{2.60\\[-2pt]
          {\footnotesize\(\pm\,0.09\)}}
        & \cellcolor{yellow!30}
          \shortstack{1.59\\[-2pt]
          {\footnotesize\(\pm\,0.04\)}}
        & \cellcolor{orange!30}
          \shortstack{1.42\\[-2pt]
          {\footnotesize\(\pm\,0.05\)}}
        & \cellcolor{orange!30}
          \shortstack{1.39\\[-2pt]
          {\footnotesize\(\pm\,0.03\)}}
        & \cellcolor{orange!30}
          \shortstack{1.34\\[-2pt]
          {\footnotesize\(\pm\,0.03\)}} \\

        \specialrule{0.25pt}{1pt}{1pt}

        W/O Residual Learning
        & \cellcolor{yellow!30}1.81
        & \cellcolor{yellow!30}2.86
        & \cellcolor{orange!30}1.52
        & \cellcolor{yellow!30}1.58
        & \cellcolor{yellow!30}1.53
        & \cellcolor{yellow!30}2.02 \\

        \specialrule{0.25pt}{1pt}{1pt}

        Full Model
        & \cellcolor{red!30}
          \shortstack{1.44\\[-2pt]{\footnotesize\(\pm\,0.07\)}}
        & \cellcolor{red!30}
          \shortstack{2.14\\[-2pt]{\footnotesize\(\pm\,0.04\)}}
        & \cellcolor{red!30}
          \shortstack{1.20\\[-2pt]{\footnotesize\(\pm\,0.04\)}}
        & \cellcolor{red!30}
          \shortstack{0.93\\[-2pt]{\footnotesize\(\pm\,0.09\)}}
        & \cellcolor{red!30}
            \shortstack{1.08\\[-2pt]{\footnotesize\(\pm\,0.13\)}}
        & \cellcolor{red!30}
         \shortstack{1.07\\[-2pt]{\footnotesize\(\pm\,0.02\)}} \\

        \bottomrule
    \end{tabular}
    \label{tab:ablation}
\end{table*}

\begin{table*}[t]

    \centering
    \caption{Comparison with constraint-enforcement alternatives on
    Dataset~1, BDLO~1. For the after-training and improvement results,
    each cell reports the mean on the first line and the standard
    deviation on the second line across three independently trained seeds.}
    \label{tab:constraint_alternatives}

    \begin{tabular}{l||ccc}
        \toprule
        & \multicolumn{3}{c}{
            Modeling Accuracy (RMSE, \(10^{-2}\,\mathrm{m}\))\(\downarrow\)
        } \\
        Method & Before Training & After Training & Improvement \\
        \midrule

        Replace with Newtonian Force
        & \cellcolor{yellow!30}4.27
        & \cellcolor{yellow!30}
          \shortstack{3.94\\[-2pt]
          {\footnotesize\(\pm\,0.26\)}}
        & \cellcolor{yellow!30}
          \shortstack{0.33\\[-2pt]
          {\footnotesize\(\pm\,0.26\)}} \\

        \specialrule{0.25pt}{1pt}{1pt}

        Without Momentum Conservation
        & \cellcolor{orange!30}3.60
        & \cellcolor{orange!30}
          \shortstack{3.17\\[-2pt]
          {\footnotesize\(\pm\,0.18\)}}
        & \cellcolor{orange!30}
          \shortstack{0.43\\[-2pt]
          {\footnotesize\(\pm\,0.18\)}} \\

        \specialrule{0.25pt}{1pt}{1pt}

        DEFT
        & \cellcolor{red!30}2.46
        & \cellcolor{red!30}
          \shortstack{1.44\\[-2pt]
          {\footnotesize\(\pm\,0.07\)}}
        & \cellcolor{red!30}
          \shortstack{1.01\\[-2pt]
          {\footnotesize\(\pm\,0.07\)}} \\

        \bottomrule
    \end{tabular}
\end{table*}

\begin{table*}[t]

    \centering
    \caption{Ablation study comparing GNN- and MLP-based residual-learning approaches. Modeling accuracy is averaged across the six BDLO/dataset evaluation settings in Table~\ref{tab:ablation}. Computational time is reported as an average across the evaluated BDLOs.}
    \begin{tabular*}{\textwidth}{@{\extracolsep{\fill}}l||cc|c@{}}
        \toprule
        Method
        & Average RMSE ($10^{-2}$ m)$\downarrow$
        & RMSE Reduction ($10^{-2}$ m)$\uparrow$
        & Average Comp. Time ($10^{-2}$ s)$\downarrow$ \\
        \hline

        W/O Residual Learning
        & \cellcolor{yellow!30}1.89
        & \cellcolor{yellow!30}0.00
        & \cellcolor{red!30}0.68 \\

        Replace GNN with MLP
        & \cellcolor{orange!30}1.80
        & \cellcolor{orange!30}0.09
        & \cellcolor{orange!30}0.71 \\

        DEFT
        & \cellcolor{red!30}1.31
        & \cellcolor{red!30}0.58
        & \cellcolor{yellow!30}0.74 \\

        \bottomrule
    \end{tabular*}
    \label{tab:residual_ablation}
\end{table*}

Table \ref{tab:ablation} highlights the impact of each DEFT contribution.
Excluding all constraints introduced in Section \ref{section:constratins} results in simulation instability, primarily because inextensibility is no longer enforced.
Similarly, removing attachment constraints allows the children branch to detach from parent branch and fall freely, leading to a substantial increase in loss.
Excluding near-rigidity constraints similarly reduces accuracy, underscoring the necessity of junction rigidity for effective dynamic propagation across branches.
 Although residual learning is used to mitigate errors, directly modeling stiff behaviors as described in Section \ref{section:constratins}, remains challenging, making the learning process highly sensitive to inputs and hampering effective gradient propagation.
Neural networks alone struggle to maintain the system within the low-dimensional manifold defined by these constraints and minor errors can accumulate during open-loop predictions, reducing residual learning effectiveness.
Despite these challenges, combination of auto-tuning the material properties and residual learning also provides the substantial improvement, illustrating DEFT’s ability to leverage real-world data for parameter identification and residual learning.
By contrast, the model that do not learn from real-world data underperforms those that do, illustrating the importance of differentiability in this framework.
\begin{figure*}[t]
    \centering
    \includegraphics[width=1\textwidth]{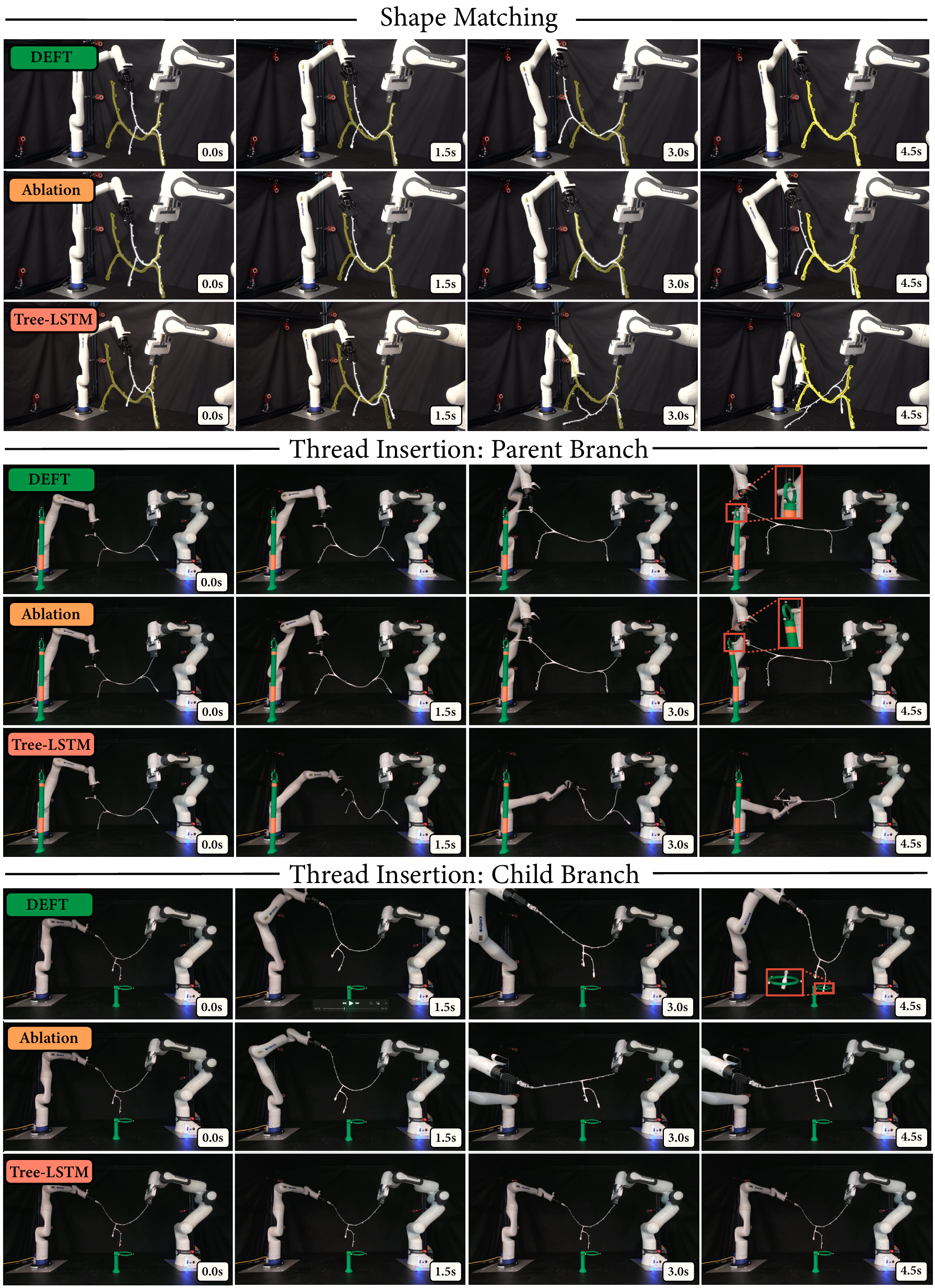}
    \caption{Visualization of planning across three manipulation tasks using DEFT, a DEFT ablation without the constraint in Theorem~\ref{thm:junctionconstraint}, and Tree-LSTM.
    The first two tasks are performed on BDLO~1, while the third task uses BDLO~5.
    In the shape matching task, the BDLO is highlighted in yellow for visibility.
    The DEFT model successfully completes all tasks, whereas both the ablation variant and Tree-LSTM fail to achieve successful execution.}
    \label{fig:planning_demo}
\end{figure*}
\subsubsection{Comparison with Constraint Enforcement Alternatives}
To further validate the importance of Theorems~\ref{theorem:constraint_attachement} and~\ref{thm:junctionconstraint} for both forward simulation and backward gradient propagation, we compare DEFT against two alternatives.
The first replaces these constraints with traditional Newtonian force formulations.
The second applies the theorems without enforcing momentum conservation and without providing derived analytical gradients, implemented using the Theseus optimizer~\citep{theseus}.
Results on Dataset~1, BDLO~1 are summarized in Table~\ref{tab:constraint_alternatives}.

DEFT outperforms both alternatives before and after training, highlighting the importance of the proposed constraints in achieving accurate forward simulation.
Moreover, DEFT shows the largest improvement after training, highlighting the necessity of momentum conservation and analytical gradients for effective parameter identification and residual learning.
In contrast, the Newtonian-force baseline yields only marginal gains, reflecting its inability to capture stiff behaviors under larger time steps and its limited support for gradient-based optimization.

\subsubsection{Comparison with Learning Architecture Alternatives}
To further evaluate the importance of the GNN for residual learning, we replace it with a fully connected MLP and evaluate both approaches across all six BDLO/dataset settings.

As shown in Table~\ref{tab:residual_ablation},
the MLP reduces the average RMSE from 1.89 to 1.80, whereas the GNN reduces it to 1.31.
This improvement requires only a modest increase in average computation time from \(0.71\times10^{-2}\)~s for the MLP to \(0.74\times10^{-2}\)~s for the GNN.
These results indicate that the topology-aware GNN captures residual BDLO dynamics more effectively than an unstructured MLP.
This improvement could be particularly important for long-horizon recursive predictions, where local errors can accumulate over time.

\begin{table}[t]

    \centering
    \caption{Ablation study comparing multi-step and single-step
    training on Dataset~1 (BDLO~1). Each entry reports the mean RMSE
    on the first line and the standard deviation on the second line
    across three independently trained seeds.}
    \small
    \begin{tabular}{l||c}
        \toprule
        & Modeling Accuracy\(\downarrow\) \\
        Method & (RMSE, \(10^{-2}\,\mathrm{m}\)) \\
        \midrule

        Single-step Training
        & \cellcolor{orange!30}
          \shortstack{2.02\\[-2pt]
          {\footnotesize\(\pm\,0.38\)}} \\

        \specialrule{0.25pt}{1pt}{1pt}

        Multi-step Training
        & \cellcolor{red!30}
          \shortstack{1.44\\[-2pt]
          {\footnotesize\(\pm\,0.07\)}} \\

        \bottomrule
    \end{tabular}
    \label{tab:ablation_appendix}
\end{table}

\begin{table*}[t]

    \centering
    \caption{Success rates of planning tasks for different methods.
    Each cell reports the number of successful trials and the corresponding
    success rate on the first line, followed by the 95\% Wilson confidence
    interval on the second line. Insertion Task~1 involves inserting the
    tip of the parent branch through the central opening of a target torus
    using BDLO~1. Insertion Task~2 involves inserting the tip of the second
    child branch through the central opening of the torus using BDLO~5.}
    \label{tab:planning}

    \small
    \begin{tabular}{l||c|c|c}
        \toprule
        & \multicolumn{3}{c}{Success Rate with 95\% Wilson CI} \\
        Methods
        & Shape Matching
        & Insertion Task 1
        & Insertion Task 2 \\
        \midrule

        Tree-LSTM
        & \shortstack{
            13/40 \;(32.5\%)\\[-2pt]
            {\footnotesize\([20.1,48.0]\%\)}
          }
        & \shortstack{
            0/40 \;(0.0\%)\\[-2pt]
            {\footnotesize\([0.0,8.8]\%\)}
          }
        & \shortstack{
            0/20 \;(0.0\%)\\[-2pt]
            {\footnotesize\([0.0,16.1]\%\)}
          } \\

        \specialrule{0.25pt}{1pt}{1pt}

        DEFORM\(^{+}\)
        & \cellcolor{yellow!30}
          \shortstack{
            0/40 \;(0.0\%)\\[-2pt]
            {\footnotesize\([0.0,8.8]\%\)}
          }
        & \cellcolor{yellow!30}
          \shortstack{
            38/40
            \;(95.0\%)\\[-2pt]
            {\footnotesize\([83.5,98.6]\%\)}
          }
        & \cellcolor{yellow!30}
          \shortstack{
            2/20 \;(10.0\%)\\[-2pt]
            {\footnotesize\([2.8,30.1]\%\)}
          } \\

        \specialrule{0.25pt}{1pt}{1pt}

        DEFT (Ablation)
        & \cellcolor{orange!30}
          \shortstack{
            0/40 \;(0.0\%)\\[-2pt]
            {\footnotesize\([0.0,8.8]\%\)}
          }
        & \cellcolor{orange!30}
          \shortstack{
            38/40
            \;(95.0\%)\\[-2pt]
            {\footnotesize\([83.5,98.6]\%\)}
          }
        & \cellcolor{orange!30}
          \shortstack{
            2/20 \;(10.0\%)\\[-2pt]
            {\footnotesize\([2.8,30.1]\%\)}
          } \\

        \specialrule{0.25pt}{1pt}{1pt}

        DEFT
        & \cellcolor{red!30}
          \shortstack{
            36/40 \;(90.0\%)\\[-2pt]
            {\footnotesize\([76.9,96.0]\%\)}
          }
        & \cellcolor{red!30}
          \shortstack{
            40/40
            \;(100.0\%)\\[-2pt]
            {\footnotesize\([91.2,100.0]\%\)}
          }
        & \cellcolor{red!30}
          \shortstack{
            16/20 \;(80.0\%)\\[-2pt]
            {\footnotesize\([58.4,91.9]\%\)}
          } \\

        \bottomrule
    \end{tabular}
\end{table*}

\subsubsection{Ablation Study on Leveraging DEFT's Differentiability}
By leveraging DEFT's differentiability, $T$ in \eqref{eq:trainingcost} can be selected as greater than 2  to better capture long-term dynamics.
Training with only single-step predictions risks overfitting to noise in the collected data, whereas multi-step training acts as a filter that smooths out such noise.
As a result, this multi-step training approach (50 steps) results in improved prediction accuracy compared to single-step training, as demonstrated in Table~\ref{tab:ablation_appendix}.
Single-step training produces both higher prediction error and greater variability across independently trained seeds, indicating greater sensitivity to noise and error accumulation during recursive prediction.
In contrast, multi-step training accounts for these accumulated deviations during training, resulting in more accurate and consistent predictions.

\subsection{Planning Results}
This section demonstrates how these BDLO models can be used for planning tasks, including shape matching and thread insertion.
We compare four approaches:
DEFT, which is our proposed method;
DEFT without the constraint described in Theorem~\ref{thm:junctionconstraint}, an ablation approach that removes rigidity at junctions and thus limits dynamic propagation across branches;
Tree-LSTM, the baseline model with performance closest to DEFT;
and DEFORM$^+$, the physics-based BDLO baseline.

We integrate DEFT into the planning framework as described in Section~\ref{section:planning}.
The results are summarized in Table~\ref{tab:planning}, and Figure~\ref{fig:planning_demo} shows one of the real-world results.
A video of these experiments can be found in the supplementary material.

\textbf{Shape Matching.}

Each trial specifies a goal configuration and a random initial condition.
A trial succeeds if each vertex reaches within \(2.5\)~cm of its target within \(10\)~seconds.
As shown in Table~\ref{tab:planning} (\textbf{left}), DEFT outperforms all the baselines.
Although DEFORM$^+$ achieves a lower modeling RMSE than Tree-LSTM on BDLO~1, it obtains \(0/40\) shape-matching successes, compared with \(13/40\) for Tree-LSTM.
This exception occurs because in some Tree-LSTM trials, the planner may have found a locally optimal trajectory that successfully reached the goal despite the larger modeling error.
Overall, the results in Table~8 show that higher modeling accuracy generally leads to higher planning success, with DEFT achieving both the lowest modeling error and the highest success rate.

\textbf{Thread Insertion.}
In this experiment, the planning algorithm must insert the end of the BDLO into a hole placed at an arbitrary location.
The initial states of the BDLO are randomized.
To demonstrate the generalization of our modeling approach and its dexterous manipulation capability, we design two thread insertion scenarios.
The first task inserts the tip of the parent branch through a target torus using BDLO~1.
The second task inserts the tip of the second child branch through the torus using BDLO~5, which requires accurate prediction of deformation propagation across the junction between the parent and child branches.
A trial is considered successful if the target tip passes through the central opening of the torus in the real-world experiment.
For the first task, we set the pre-insertion offset $\delta_{\mathrm{pre}} = 5$ cm and the insertion depth $\delta_{\mathrm{ins}} = -5$ cm.
In this task, the target hole has a radius of $2.5$ cm.
For the second task, we set the pre-insertion offset $\delta_{\mathrm{pre}} = 5$ cm and the insertion depth $\delta_{\mathrm{ins}} = -2.5$ cm.
In this case, the target hole has a radius of \(5\)~cm.
As shown in Table~\ref{tab:planning} (\textbf{middle}, \textbf{right}), DEFT outperforms all baselines.
Note that DEFORM\ensuremath{^+} and the ablated DEFT achieve performance comparable to DEFT when performing thread insertion of the parent branch, as they can still accurately predict the dynamics of the parent branch.
However, they fail to complete most of the child branch insertion tasks.
DEFT succeeds in \(16/20\) trials for Insertion Task~2. Among the four failures, some result from prediction errors that cause the target branch to miss the hole, while others result from undesirable inverse-kinematics solutions generated by the robot-arm controller. Representative failure cases are included in the supplementary video.
\section{Discussion}
\label{sec:discussion}

The presented results highlight DEFT's potential as a foundation for autonomous BDLO manipulation in robotics.
However, DEFT still has limitations related to computational scalability, sensitivity and robustness, and applicability across different cable materials and geometries.

\subsection{Limitations}

\noindent\textbf{Computational scalability.}
As shown in Figure~\ref{fig:computational_speed3}, the parallel implementation scales more favorably than the sequential implementation, although computation time still increases as more branches are added.
Two factors contribute to this increase.
First, enforcing the constraints in Figure~\ref{fig:deft_contribution_diagram} requires more iterations to converge as the number of branches increases.
Second, the cost grows with the number of vertices because Algorithm~\ref{alg:constraints enforcement} enforces constraints sequentially, edge by edge.
These results identify constraint enforcement as the primary opportunity for further improving computational scalability.
A projective-dynamics-style approach could enforce these constraints in parallel and improve convergence.
\medskip

\noindent\textbf{Sensitivity and robustness.}
The experiments demonstrate accurate long-horizon open-loop predictions under the evaluated conditions.
However, because DEFT performs recursive prediction, errors in the initial state, control inputs, physical parameters, or residual dynamics may accumulate over time.
A systematic evaluation of sensitivity to such perturbations and robustness to unmodeled disturbances, together with a finer-grained analysis of how prediction errors accumulate over different horizons and affect task success, remains an important direction for future work.

\noindent\textbf{Applicability to different cable materials and geometries.}
We also note the scope of DEFT's generalization.
The model formulation, including the physics-based dynamics, constraint enforcement, and residual GNN architecture, is topology-agnostic and
can accommodate BDLOs with variable numbers of branches, vertices, and junctions without architectural redesign.
The model parameters, by contrast, are fitted per BDLO from data: the bending stiffness, twisting stiffness, and mass governing each BDLO's deformation depend on its material composition, and the residual GNN compensates for integration errors specific to that BDLO's dynamics.
Transferring to a new material therefore currently requires retraining or fine-tuning.
DEFT's differentiability makes this practical, since analytical gradients allow the parameters to be fitted using modest amounts of data.
This represents formulation-level generalization rather than zero-shot transfer of fitted physical or learned parameters between BDLOs.

Zero-shot transfer of the fitted parameters to cables with substantially different materials or geometries has not yet been systematically evaluated.
Because these differences affect both the physical parameters and residual dynamics, transferring to a new BDLO may require parameter reidentification or fine-tuning, consistent with prior work on model adaptation and online parameter estimation~\citep{adaptationdynamics,onlineadapt}.

\subsection{Broader Applicability and Future Directions}

Several opportunities remain open for exploration.
First, for the insertion tasks considered in this work, we assume that the location and geometry of the target hole are known in advance.
In unstructured or partially observed environments, this assumption may not hold.
Incorporating 3D scene understanding methods, such as Gaussian Splatting~\citep{gaussiansplatting, splanning, gsposeest}, could help autonomously detect or refine the location of target features, thereby broadening the system's applicability.
Second, we plan to explicitly incorporate gripper orientation into the dynamics, building on prior insights~\citep{diminishingridgidity} that orientation significantly affects BDLO kinematics.
Third, we aim to extend DEFT to contact-rich tasks by modeling contact forces and friction through existing frameworks~\citep{DER_contact, contactnets, physicslearn2}, enabling robust performance in scenarios with collisions and tight clearances.

Finally, the shear-free rod formulation underlying DEFT is appropriate for thin wires but neglects shear effects that may become significant for thicker rods, strong contact regimes, or filaments with large cross-sectional deformations.
Extending DEFT to a shear-inclusive Cosserat formulation following~\citet{shearDER} would broaden its applicability to such regimes, including soft robotic continuum structures.
The framework's constraint enforcement and residual learning would carry over directly, while the rod model itself, the potential energy, and the analytical gradient in Theorem~\ref{thm:potential_energy_gradient} would require extension.

Beyond BDLOs, the physics principles embedded in DEFT are transferable to other deformable objects with related topologies.
For instance, cloth~\citep{LargeStepSimulation, PBD, diffcloth}, deformable suction grippers~\citep{suctiongripper, 10802323}, and soft robotic structures~\citep{softrobot1} could be modeled within the same differentiable framework by introducing appropriate sets of constraints or by replacing the internal force formulation with alternative models.
Due to the focused scope of this work, comprehensively addressing these future directions is deferred to future research.
\section{Conclusion}
This paper introduced DEFT, a method that integrates residual learning within a novel differentiable simulator for BDLOs.
DEFT enables accurate and real-time prediction of BDLO dynamics over extended horizons under dynamic manipulation.
Through a comprehensive set of experiments, we evaluated its accuracy, computational efficiency, component contributions, and utility for manipulation tasks.
Compared to state-of-the-art approaches, DEFT achieves higher accuracy while remaining sample-efficient and computationally fast.
We further demonstrated its integration with planning and control frameworks for 3D shape matching and thread insertion, where DEFT successfully completed more complex real-world manipulation tasks than competing baselines.

In the future, extending DEFT along the identified research directions may pave the way toward a more general and autonomous framework for wireharness manipulation, capable of operating in unstructured and contact-rich environments.
We believe DEFT could provide a strong foundation for advancing autonomous wireharness assembly and opens new opportunities for broader robotic applications involving complex deformable systems.

\begin{dci}
The author(s) declared no potential conflicts of interest with respect to the research, authorship, and/or publication of this article.
\end{dci}

\begin{funding}
This work was supported by Ford Motor Company.
\end{funding}

\bibliographystyle{SageH}
\bibliography{references}

\clearpage
\appendix
\section*{Appendix}
\section{Visualization}
\label{app_sec:modeling_vis}
\begin{figure}[h]
    \centering
    \makebox[\textwidth][c]{%
        \includegraphics[width=0.833\textwidth]{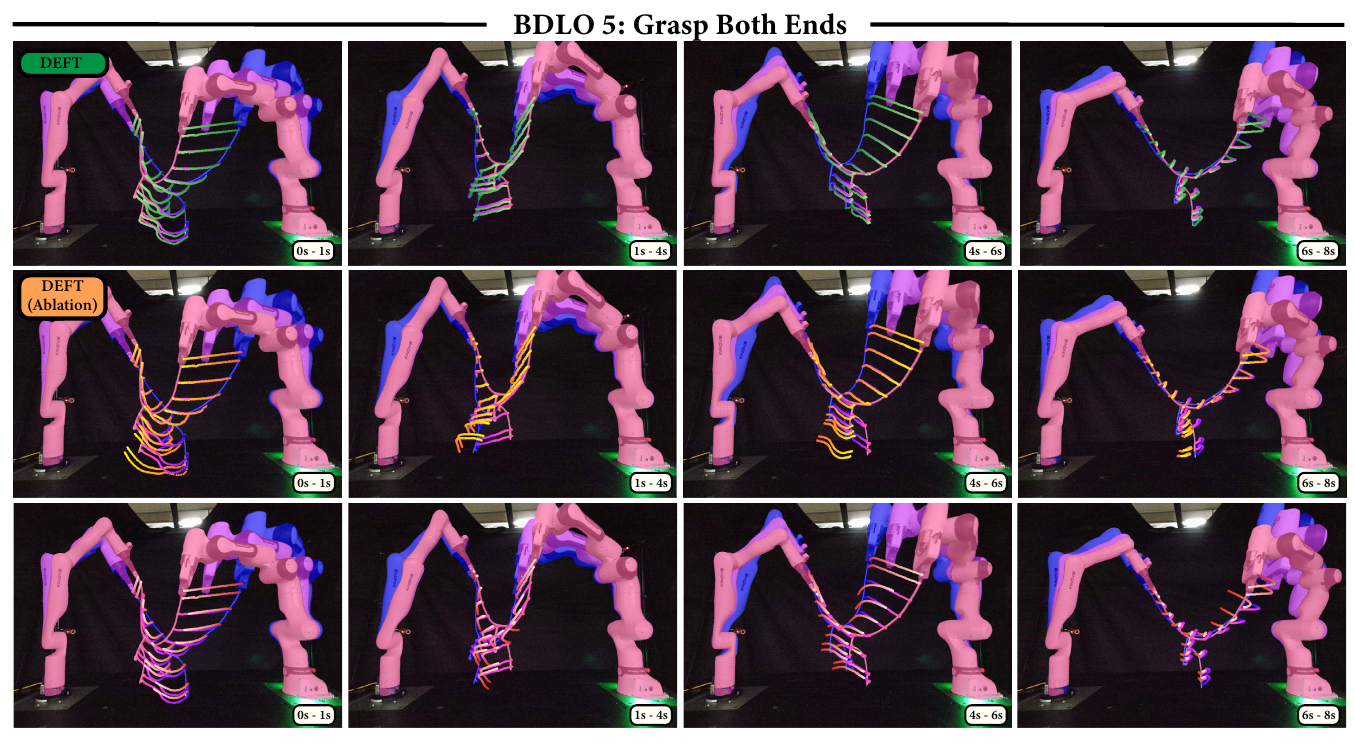}
    }
    \makebox[\textwidth][c]{%
        \begin{minipage}{0.833\textwidth}
            \caption{Visualization of the predicted trajectories using DEFT, a DEFT ablation that leaves out the constraint described in Theorem~\ref{thm:junctionconstraint}, and Tree-LSTM on predicting the behavior of BDLO 5.}
            \label{app_fig:modeling_vis}
        \end{minipage}
    }
\end{figure}
\section{Training Details}

\begin{table}[!ht]

    \centering
    \caption{Architecture and training configuration of the residual GNN in DEFT.}
    \label{tab:mlp_training_details}
    \small
    \begin{tabular}{
        @{}
        p{0.35\columnwidth}
        p{0.58\columnwidth}
        @{}
    }
        \toprule
        Configuration & Value \\
        \midrule

        MLP architecture
        & \(16\!-\!256\!-\!256\!-\!256\!-\!3\)
          (3 hidden layers) \\

        Hidden-layer activation
        & LeakyReLU after each hidden layer
          (negative slope \(0.2\)) \\

        Output activation
        & None (linear output) \\

        Loss function
        & Mean squared error \\

        Optimizer
        & Adam (\(\epsilon_{\mathrm{Adam}}=10^{-8}\)) \\

        GNN learning rate
        & \(10^{-4}\) \\

        Batch size
        & 32 \\

        Training epochs
        & 100 \\

        Training horizon
        & 50 time steps (\(0.5\,\mathrm{s}\)) \\

        \bottomrule
    \end{tabular}
\end{table}
\section{Empirical Constraint Satisfaction}
\label{appendix:constraint_satisfaction}

\begin{figure*}[t]

    \centering
    \includegraphics[width=0.82\textwidth]{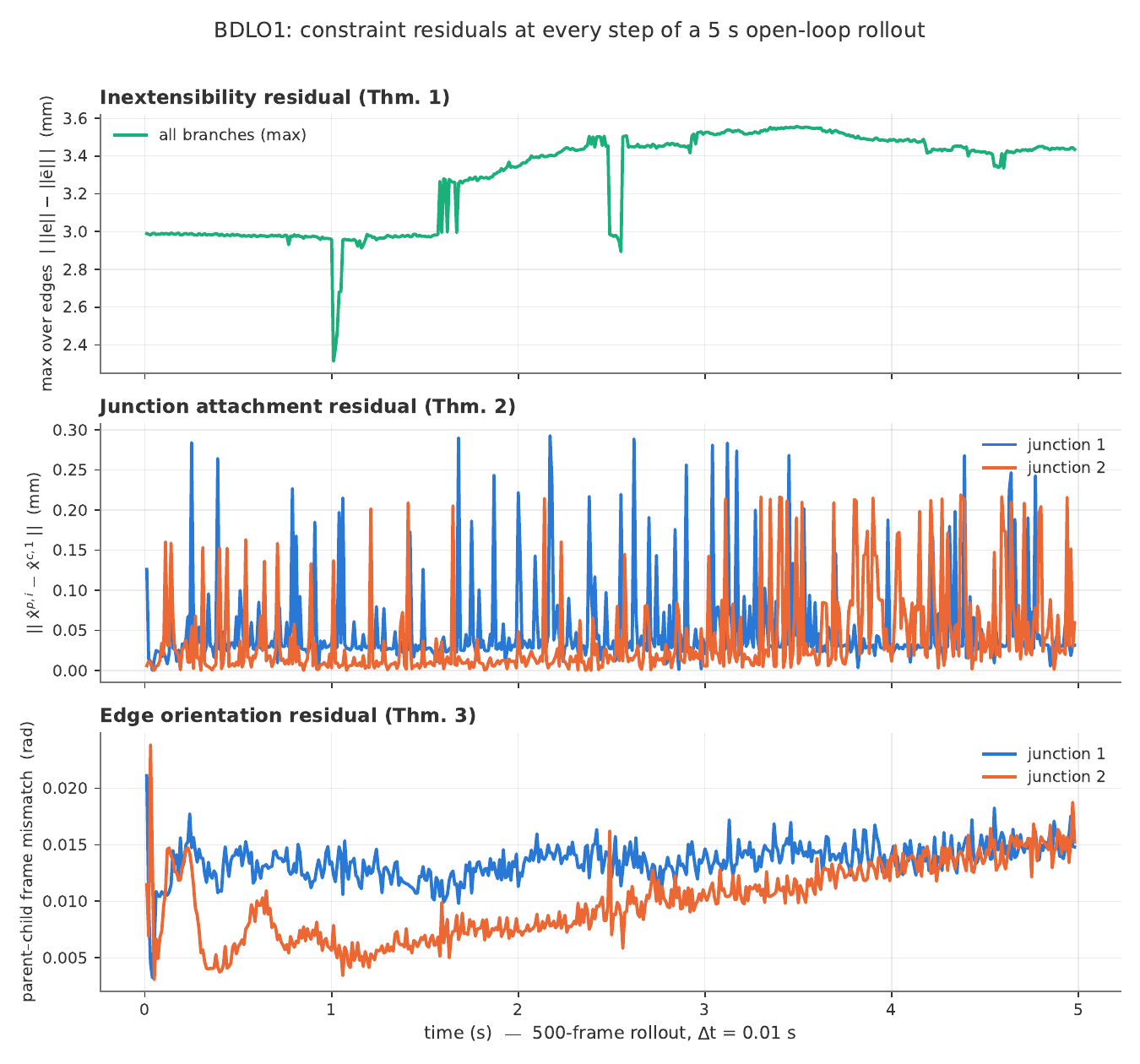}
    \caption{Constraint residuals at every simulation step of a \(5\)~s open-loop rollout on BDLO~1. The panels report the maximum inextensibility residual across all branches, the junction-attachment residuals, and the parent--child orientation residuals. The corresponding squared constraint values remain below a tight user defined threshold throughout the rollout.}
    \label{fig:constraint_satisfaction}
\end{figure*}

Figure~\ref{fig:constraint_satisfaction} reports the constraint residuals at every simulation step of a \(5\)~s, \(500\)-step open-loop rollout on BDLO~1.
The corresponding squared residuals remain below the stopping threshold throughout the rollout, demonstrating stable empirical convergence under the evaluated conditions.

\end{document}